\documentclass[sigconf]{acmart}

\usepackage{bm}
\usepackage{amsmath, amsfonts}
\usepackage{float}
\usepackage[ruled,vlined,linesnumbered]{algorithm2e}
\usepackage{algpseudocode}
\usepackage{leftidx}
\usepackage{url}
\usepackage{graphicx}
\usepackage{tabularx, multirow, multicol}
\usepackage{color}
\usepackage{soul}
\usepackage{subfigure}
\usepackage{caption}
\usepackage{float}
\usepackage{balance}

\newcolumntype{P}{>{\centering\arraybackslash}m{0.05\linewidth}}
\newcolumntype{Q}{>{\centering\arraybackslash}m{0.09\linewidth}}
\newcolumntype{R}{>{\raggedleft\arraybackslash}m{0.095\linewidth}}
\newcolumntype{S}{>{\centering\arraybackslash}m{0.085\linewidth}}
\newcolumntype{T}{>{\centering\arraybackslash}m{0.135\linewidth}}

\AtBeginDocument{%
  \providecommand\BibTeX{{%
    \normalfont B\kern-0.5em{\scshape i\kern-0.25em b}\kern-0.8em\TeX}}}

\copyrightyear{2021}
\acmYear{2021}
\setcopyright{acmcopyright}\acmConference[SIGIR '21]{Proceedings of the 44th International ACM SIGIR Conference on Research and Development in Information Retrieval}{July 11--15, 2021}{Virtual Event, Canada}
\acmBooktitle{Proceedings of the 44th International ACM SIGIR Conference on Research and Development in Information Retrieval (SIGIR '21), July 11--15, 2021, Virtual Event, Canada}
\acmPrice{15.00}
\acmDOI{10.1145/3404835.3462935}
\acmISBN{978-1-4503-8037-9/21/07}

\newcommand{\proposed}{BUIR\xspace}
\newcommand{\proposedid}{BUIR\textsubscript{id}\xspace}
\newcommand{\proposednb}{BUIR\textsubscript{nb}\xspace}

\newcommand{\bpr}{BPR\xspace}
\newcommand{\cml}{CML\xspace}
\newcommand{\sml}{SML\xspace}
\newcommand{\neumf}{NeuMF\xspace}

\newcommand{\vae}{M-VAE\xspace}

\newcommand{\cfgan}{CFGAN\xspace}

\newcommand{\ngcf}{NGCF\xspace}
\newcommand{\lgcn}{LGCN\xspace}

\newcommand{\ciao}{Ciao\xspace}
\newcommand{\culike}{CiteULike\xspace}

\newcommand{\foursq}{FourSquare\xspace}

\newcommand{\predfn}[1]{q_\theta\left(#1\right)}

\newcommand{\ouser}{f_\theta({u})}
\newcommand{\tuser}{f_\xi({u})}
\newcommand{\oitem}{f_\theta({v})}
\newcommand{\titem}{f_\xi({v})}

\newcommand{\userset}{\mathcal{V}_u}
\newcommand{\itemset}{\mathcal{U}_v}
\newcommand{\ouserset}{f_\theta(u, \mathcal{V}_u)}
\newcommand{\oitemset}{f_\theta(v, \mathcal{U}_v)}

\newcommand{\smallsection}[1]{{\vspace{0.05in} \noindent \bf {#1.\hspace{5pt}}}}

\begin{document}
\fancyhead{}

\title{Bootstrapping User and Item Representations for~One-Class~Collaborative~Filtering}


\author{Dongha Lee$^1$, SeongKu Kang$^1$, Hyunjun Ju$^1$, Chanyoung Park$^2$, Hwanjo Yu$^1$}
\authornote{corresponding author}
\affiliation{%
  \institution{$^1$Pohang University of Science and Technology (POSTECH), South Korea}
  \institution{$^2$Korea Advanced Institute of Science and Technology (KAIST), South Korea}
  \{dongha.lee, seongku, hyunjunju, hwanjoyu\}@postech.ac.kr, cy.park@kaist.ac.kr
}



\begin{abstract}
The goal of one-class collaborative filtering (OCCF) is to identify the user-item pairs that are positively-related but have not been interacted yet, where only a small portion of positive user-item interactions (e.g., users' implicit feedback) are observed.
For discriminative modeling between positive and negative interactions, most previous work relied on negative sampling to some extent, which refers to considering unobserved user-item pairs as negative, as actual negative ones are unknown.
However, the negative sampling scheme has critical limitations because it may choose ``positive but unobserved'' pairs as negative.
This paper proposes a novel OCCF framework, named as \proposed, which does not require negative sampling.
To make the representations of positively-related users and items similar to each other while avoiding a collapsed solution, 
\proposed adopts two distinct encoder networks that learn from each other;
the first encoder is trained to predict the output of the second encoder as its target, while the second encoder provides the consistent targets by slowly approximating the first encoder.
In addition, \proposed effectively alleviates the data sparsity issue of OCCF, by applying stochastic data augmentation to encoder inputs.
Based on the neighborhood information of users and items,
\proposed randomly generates the augmented views of each positive interaction each time it encodes, then further trains the model by this self-supervision.
Our extensive experiments demonstrate that \proposed consistently and significantly outperforms all baseline methods by a large margin especially for much sparse datasets in which any assumptions about negative interactions are less valid.

\end{abstract}



\maketitle

\section{Introduction}
\label{sec:intro}

Over the past decade, one-class collaborative filtering (OCCF) problems~\cite{pan2008one, hu2008collaborative} have been extensively researched to accurately infer a user's preferred items, particularly for the recommender systems where only the users' implicit feedback on items are observed (e.g., click, purchase, or browsing history).
This problem has remained challenging
due to an extreme sparseness of such implicit feedback (i.e., most users have interacted with only a few items among numerous items), and also the non-existence of the negative labels for user-item interactions
(i.e., observed feedback is expressions of positive interactions).
Precisely, the goal of OCCF is to identify the most likely positive user-item interactions among a huge amount of unobserved interactions, by using only a small number of observed (positively-labeled) interactions.

The most dominant approach to the OCCF problem is discriminative modeling~\cite{rendle2009bpr, hsieh2017collaborative, he2017neural, li2020symmetric, kim2019dual, wang2019neural}, which explicitly aims to distinguish positive user-item interactions from the negative counterparts.
They define the \textit{interaction score} indicating how likely each user interacts with each item, based on the similarity (e.g., inner product) between the representation of a user and an item.
From matrix factorization~\cite{hu2008collaborative, rendle2009bpr} to deep neural networks~\cite{he2017neural, wang2019neural}, a variety of techniques have been studied to effectively model this score.
Then, they optimize the scores by using the pointwise prediction loss~\cite{hu2008collaborative, he2017neural} or the pairwise ranking loss~\cite{rendle2009bpr, hsieh2017collaborative} to discriminate between positive and negative interactions.

However, since the negative interactions are not available in the OCCF problem, previous discriminative methods assume that all unobserved interactions are negative.
In other words, for each user, the items that have not been interacted yet are regarded to be less preferred to positive items.
In this sense, they either use all unobserved user-item interactions as negative or adopt a \textit{negative sampling}, which randomly samples unobserved user-item interactions in a stochastic manner to alleviate the computational burden.
For better recommendation performance and faster convergence, advanced negative sampling strategies \cite{rendle2014improving, ding2019sampler} are also proposed to sample from non-uniform distributions.

Nevertheless, the negative sampling approach has critical limitations in the following aspects.
First, the underlying assumption about negative interactions becomes less valid as user-item interactions get sparser.
This is because as fewer positive interactions are observed, the number of "positive but unobserved" interactions increases, which consequently makes it even harder to sample correct negative ones.
Such uncertainty of supervision eventually degrades the performance for top-$K$ recommendation.
Second, the convergence speed and the final performance depend on the specific choice of distributions for negative sampling.
For example, sampling negative pairs from a non-uniform distribution~\cite{rendle2014improving, ding2019sampler} (e.g., the multinomial distribution which models the probability of each interaction being actually negative) can improve the final performance, but inevitably incurs high computational costs, especially when a lot of users and items should be considered.

As a solution to the aforementioned limitations, this paper proposes a novel OCCF framework, named as \proposed, which does not require the negative sampling at all for training the model.
The main idea is, given a positive user-item interaction ($u$, $v$), to make representations for $u$ and $v$ similar to each other, in order to encode the preference information into the representations.
However, a naive end-to-end learning framework that guides positive user-item pairs to be similar to each other without any negative supervision can easily converge to a \textit{collapsed solution} -- the encoder network outputs the same representations for all the users and items.

We argue that the above collapsed solution is incurred by the simultaneous optimization of $u$ and $v$ within the end-to-end learning framework of a single encoder. 
Hence, we instead adopt the student-teacher-like network~\cite{tarvainen2017mean, grill2020bootstrap} in which only the student's output $u$ (and $v$) is optimized to predict the target $v$ (and $u$) presented by the teacher.
Specifically, \proposed directly bootstraps\footnote{In this paper, the term ``bootstrapping'' is not used in the statistical meaning, but in the idiomatic meaning~\cite{grill2020bootstrap}. Strictly speaking, it refers to using estimated values (i.e., the output of networks) for estimating its target values, which serve as supervision for the update. For instance, semi-supervised learning based on predicted pseudo-labels~\cite{tarvainen2017mean} also can be thought as a bootstrapping method.}
the representations of users and items by employing two distinct encoder networks, referred to as \textit{online encoder} and \textit{target encoder}. 
The high-level idea is training only the online encoder for the prediction task between $u$ and $v$, where the target for its prediction is provided by the target encoder.
That is, the online encoder is optimized so that its user (and item) vectors get closer to the item (and user) vectors computed by the target encoder.
At the same time, the target encoder is updated based on momentum-based moving average~\cite{tarvainen2017mean, he2020momentum, grill2020bootstrap} to slowly approximate the online encoder, which encourages to provide enhanced representations as the target for the online encoder.
By doing so, the online encoder can capture the positive relationship between $u$ and $v$ into the representations, while preventing the model from collapsing to the trivial solution without explicitly using any negative interactions for the optimization.

Furthermore, we introduce a stochastic data augmentation technique to relieve the data sparsity problem in our framework.
Motivated by the recent success of self-supervised learning in various domains~\cite{chen2020simple,devlin2019bert}, we exploit \textit{augmented views} of an input interaction, which are generated based on the neighborhood information of each user and item (i.e., the set of the items interacted with a user, and the users interacted with an item).
The stochastic augmentation is applied to positive user-item pairs when they are passed to the encoder, so as to produce the different views of the pairs.
To be precise, by making our encoder use a random subset of a user's (and item's) neighbors for the input features, it produces a similar effect to increasing the number of positive pairs from the data itself without any human intervention.
In the end, \proposed is allowed to learn various views of each positive user-item pair.

Our extensive evaluation on real-world implicit feedback datasets shows that \proposed consistently performs the best for top-$K$ recommendation among a wide range of OCCF methods.
In particular, the performance improvement becomes more significant in sparser datasets, with the help of utilizing augmented views of positive interactions as well as eliminating the effect of uncertain negative interactions.
In addition, comparison results on a downstream task, which classifies the items into their category, support that \proposed learns more effective representations than other OCCF baselines.

%
%
%

\section{Related Work}
\label{sec:related}

\subsection{One-Class Collaborative Filtering}
\label{subsec:occf}
One-class collaborative filtering (OCCF) was firstly introduced to handle the real-world recommendation scenario where only positive user-item interaction can be labeled~\cite{pan2008one, hu2008collaborative} as a form of users' implicit feedback on items.
That is, only the set of positive user-item pairs, denoted by $\mathcal{R}$, is given for training the model.
The main challenge of OCCF is to find out the most likely positive interactions among a large number of unobserved user-item pairs in which both positive and negative interactions are mixed together.
To handle the absence of negatively-labeled interactions, most existing methods have either treated all unobserved user-item pairs as negative, or sampled some of them~\cite{he2017neural}, assuming that the items that have not been interacted yet are less preferred to positive items.

To be specific, discriminative methods~\cite{rendle2009bpr, wu2016collaborative, hsieh2017collaborative, he2017neural, li2020symmetric, kim2019dual, wang2019neural} train their model so that it can differentiate the scores between positive and negative interactions.
Pairwise learning, which is the most popular approach to personalized ranking, explicitly utilizes the pairs of positive and negative interactions for training.
Formally, the pairwise ranking loss optimizes the similarity for a positive interaction to become larger than that for a negative one as follows.
\begin{equation}
    \begin{split}
        \mathcal{L}=-\sum_{(u, v^p, v^n)\in \mathcal{O}} \phi(sim(u, v^p) > sim(u, v^n)),
    \end{split}
\end{equation}
where $\mathcal{O}=\{(u, v^p, v^n)|(u, v^p)\in \mathcal{R}, (u, v^n) \notin \mathcal{R}\}$, and $\phi$ is a scoring function to facilitate the optimization.
For example, Bayesian personalized ranking~\cite{he2016vbpr, kim2019dual, wang2019neural} defines the similarity of a user and an item by the inner product of their representations, and collaborative metric learning~\cite{hsieh2017collaborative, park2018collaborative, li2020symmetric} directly learns the latent space by modeling their similarity as the Euclidean distance.
However, all these methods obtain the negative interactions from unobserved user-item pairs, thus the convergence speed and final performance largely depend on the negative sampling distribution~\cite{rendle2014improving}.


On the other hand, generative methods~\cite{liang2018variational, wang2017irgan, chae2018cfgan, liu2019deep} aim to learn the underlying latent distribution of users, usually represented by binary vectors indicating their interacted items.
They employ the architecture of variational autoencoder (VAE)~\cite{liang2018variational} or generative adversarial networks (GAN)~\cite{wang2017irgan, chae2018cfgan, liu2019deep}, in order to infer the users' preference on each item based on the reconstructed (or generated) user vectors.
Rather than exploiting the negative sampling, most of the generative methods implicitly assume that all unobserved user-item pairs are negative in that they learn the partially-observed binary vectors as their inputs.
We remark that this assumption is not strictly valid, which eventually leads to limited performance.

\subsection{Self-supervised Contrastive Learning}
\label{subsec:contrastive}
Recently, a self-supervised learning approach has achieved a great success in computer vision and natural language understanding~\cite{chen2020simple, devlin2019bert, he2020momentum}.
Most of them basically adopt contrastive learning, which optimizes the representations of positively-related (similar) instances to be close, while those of negatively-related (dissimilar) ones far from each other.
Given an unlabeled dataset $\mathcal{D}=\{x_1,\ldots,x_N\}$, positive pairs for each instance $(x, x{}^p)$ is usually obtained from the data itself (i.e., data augmentation), such as geometric transformations on a target image.
Note that it does not require any human annotations or additional labels, thus this approach falls into the category of self-supervised learning.
The noise contrastive estimator (NCE) loss~\cite{gutmann2010noise, oord2018representation} mainly used for contrastive learning is defined by using all the other instances except for $x$ as negative:
\begin{equation}
    \mathcal{L} = -\sum_{x\in\mathcal{D}}\log\frac{\exp(sim(x, x{}^p))}{\exp(sim(x, x{}^p))+\sum_{x{}^n\in \mathcal{D}\backslash\{x\}}\exp(sim(x, x{}^n))}.
\end{equation}
In case of large-scale datasets, the predefined number of negative instances can be selected (i.e., negative sampling).
For contrastive learning, negative pairs must be considered for its optimization so as to prevent the representations of all instances from being similar, which is known as the problem of collapsed solutions.

Pointing out that the contrastive methods need to carefully treat the negative instances during the training for effectiveness and efficiency, the most recent work proposed a bootstrapping-based self-supervised learning framework~\cite{grill2020bootstrap, chen2020exploring}, which is capable of avoiding the collapsed solution without the help of negative instances.
Inspired by bootstrapping methods in deep reinforcement learning~\cite{mnih2015human, mnih2016asynchronous}, it directly bootstraps the representation of images by using two neural networks that iteratively learn from each other.
This approach achieves the state-of-the-art performance for various downstream tasks in computer vision, and also shows better robustness to the choice of data augmentations used for self-supervision.

\section{\proposed: Proposed Framework}
\label{sec:method}
In this section, we present our OCCF framework, named as \proposed, which learns the representations of users and items without any assumptions about negative interactions.
We first describe the overall learning process with a simple encoder that takes the user-id and item-id as its input (Section~\ref{subsec:buir}) and how to infer the interaction score using the representations (Section~\ref{subsec:scoring}). 
We also introduce a stochastic data augmentation technique with an extended encoder to further exploit the neighborhood information (Section~\ref{subsec:nbaug}).

\subsection{Problem Formulation}
\label{subsec:problem}
Let $\mathcal{U}=\{u_1, \ldots, u_M\}$ and $\mathcal{V}=\{v_1, \ldots, v_N\}$ be the set of $M$ users and $N$ items, respectively.
Given a set of observed user-item interactions $\mathcal{R}=\{(u, v)| \text{user $u$ is interacted}$ $\text{with item $v$} \}$, the goal of OCCF is to obtain the interaction (or preference) score $s(u, v)\in\mathbb{R}$ indicating how likely the user $u$ interacts with (or prefers to) the item $v$.
Based on the interaction scores, we can recommend $K$ items with the highest scores for each user, called as top-$K$ recommendation.
To define the interaction score by using the representations of users and items, we focus on training the encoder network that maps each user and item into a low-dimensional latent space where the users' preferences on the items are effectively captured.

\subsection{Bootstrapping the Representations}
\label{subsec:buir}
Let $f$ be the encoder network to produce the representations of users and items.
The simplest architecture of the encoder is a single embedding layer (i.e., embedding matrix); 
this maps each user-id (or item-id) into a $D$-dimensional embedding vector that represents the latent factors of the user (or item).
Specifically, each encoder consists of a user encoder and an item encoder, and they take a one-hot vector indicating the user-id and item-id as their input.

\begin{figure}[t]
    \centering
    \includegraphics[width=\linewidth]{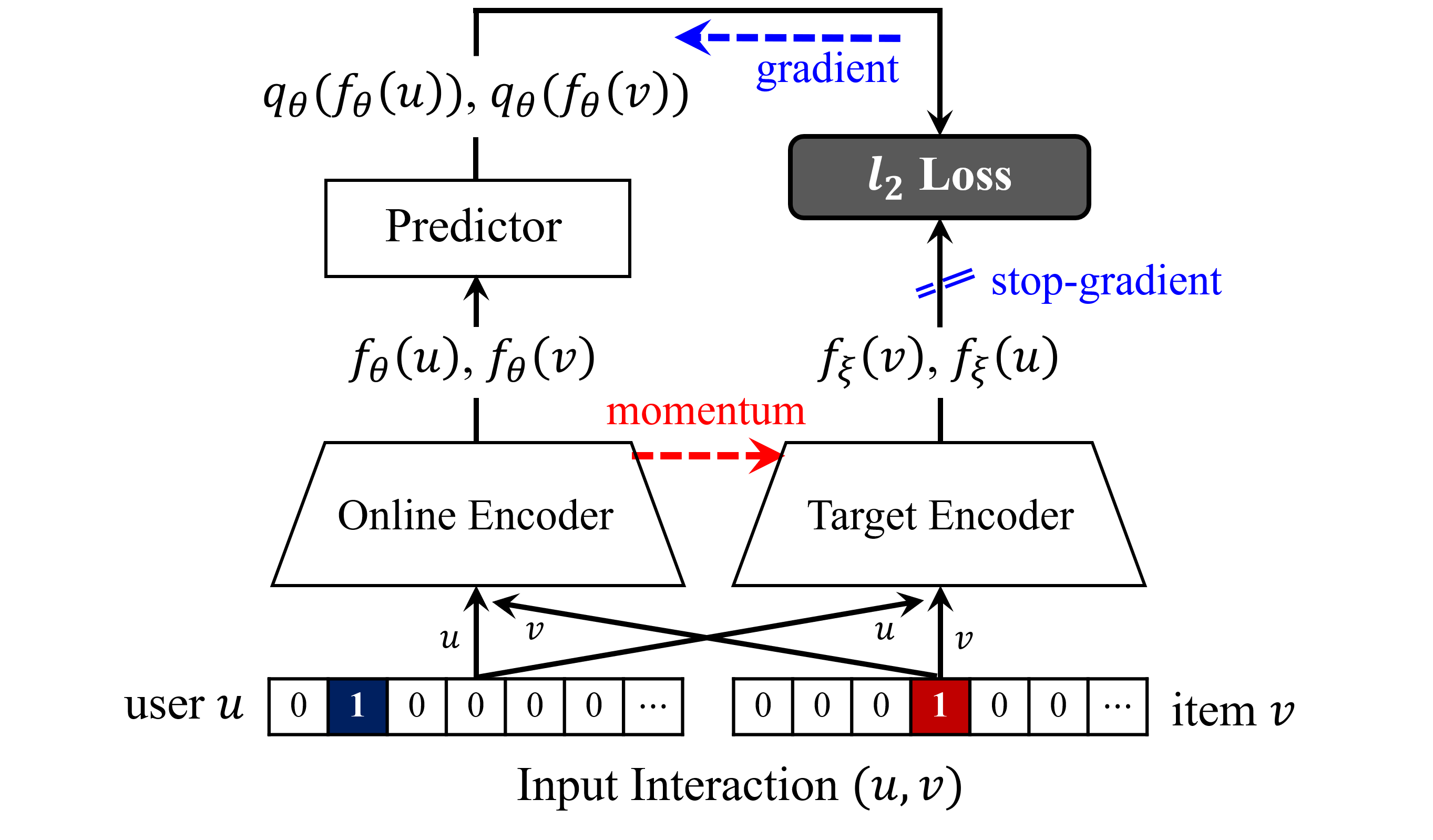}
    \caption{The overall \proposed framework.}
    \label{fig:framework}
\end{figure}

\proposed makes use of two distinct encoder networks that have the same structure: \textit{online encoder} $f_\theta$ and \textit{target encoder} $f_\xi$.
They are parameterized by $\theta$ and $\xi$, respectively.
The key idea of \proposed is to train the online encoder by using outputs of the target encoder as its target, while gradually improving the target encoder as well.
The main difference of \proposed from existing end-to-end learning frameworks is that $f_\theta$ and $f_\xi$ are updated in different ways.
The online encoder is trained to minimize the error between its output and the target, whereas the target network is slowly updated based on the momentum update~\cite{he2020momentum} so as to keep its output consistent.

To be precise, for each observed interaction $(u, v)\in\mathcal{R}$, the \proposed loss is defined based on the mean squared error of the prediction against each other (i.e., representations of $u$ and $v$) using the predictor $q_\theta:\mathbb{R}^D\rightarrow\mathbb{R}^D$ on top of the online encoder.
It includes two error terms: 
one is for updating the \textit{online} user vector $\ouser$ to accurately predict the \textit{target} item vector $\titem$, and the other is for updating the \textit{online} item vector $\oitem$ to make its prediction as the \textit{target} user vector $\tuser$.
Finally, the loss is described as follows:
\begin{equation}
\label{eq:loss}
\begin{split}
    \mathcal{L}_{\theta, \xi} (u, v) &= {l_2}\left[ \predfn{\ouser}, \titem \right] +  {l_2}\left[ \predfn{\oitem}, \tuser \right] \\
    &\approx -\frac{\predfn{\ouser}^\top \titem}{\lVert\predfn{\ouser}\rVert_2 \lVert\titem\rVert_2} -\frac{\predfn{\oitem}^\top \tuser}{\lVert\predfn{\oitem}\rVert_2 \lVert\tuser\rVert_2},
\end{split}
\end{equation}
where $l_2[\mathbf{x}, \mathbf{y}]$ is the $l_2$ distance between two normalized vectors $\overline{\mathbf{x}}$ and $\overline{\mathbf{y}}$;
i.e., $\overline{\mathbf{x}}=\mathbf{x}/\Vert\mathbf{x}\Vert_2$ and $\overline{\mathbf{y}}=\mathbf{y}/\Vert\mathbf{y}\Vert_2$.
Since the mean squared errors between two normalized vectors are equivalent to the negative value of their inner product (Equation~\eqref{eq:loss}), we simply use the inner product for the optimization.
Note that \proposed updates $\ouser$ to be similar with $\titem$ instead of $\oitem$ through the predictor, and vice versa.
This is because directly reducing the error between $\ouser$ and $\oitem$ leads to the collapsed representations when negative interactions are not considered at all for training the encoder.


To sum up, the parameters of the online encoder and target encoder are optimized by
\begin{equation}
\label{eq:opt}
\begin{split}
    \theta &\leftarrow \theta - \eta \cdot \nabla_\theta \mathcal{L}_{\theta, \xi} \\
    \xi &\leftarrow \tau\cdot\xi + (1-\tau)\cdot\theta.
\end{split}
\end{equation}
$\eta$ is the learning rate for stochastic optimization, and $\tau\in[0, 1]$ is a momentum coefficient (also called as target decay) for momentum-based moving average.
The online encoder $f_\theta$ (and the predictor $q_\theta$) is effectively optimized by the gradients back-propagated from the loss (Equation~\eqref{eq:loss}), while the target encoder $f_\xi$ is updated as the moving average of the online encoder. 
By taking a large value of $\tau$, the target encoder slowly approximates the online encoder.
This momentum-based update makes $\xi$ evolve more slowly than $\theta$, which enables to \textit{bootstrap} the representations by providing enhanced but consistent targets to the online encoders~\cite{he2020momentum, grill2020bootstrap}.
Figure~\ref{fig:framework} illustrates the overall framework of \proposed with the simple one-hot encoders.

\smallsection{Bypassing the collapsed solution}
It is obvious that the loss in Equation~\eqref{eq:loss} admits the collapsed solution with respect to $\theta$ and $\xi$, which means both the encoders generate the same representations for all users and items.
For this reason, the conventional end-to-end learning strategy, which optimizes both $f_\theta$ and $f_\xi$ to minimize the loss (i.e., cross-prediction error), may easily lead to such collapsed solution.
In contrast, our proposed framework updates each of the encoders in different ways.
From Equation~\eqref{eq:opt}, the online encoder is optimized to minimize the loss, while the target encoder is updated to slowly approximate the online encoder.
That is, the direction of updating the target encoder ($\theta-\xi$) totally differs from that of updating the online encoder ($-\nabla_\theta\mathcal{L}_{\theta,\xi}$),
and this effectively keeps both the encoders from converging to the collapsed solution.
Several recent work on bootstrapping-based representation learning~\cite{grill2020bootstrap, chen2020exploring} empirically demonstrated that this kind of dynamics (i.e., updating two networks differently) allows to avoid the collapsed solution without any explicit term to prevent it.

\subsection{Top-K Preferred Item Prediction}
\label{subsec:scoring}
To retrieve $K$ most preferred items for each user (i.e., user-item interactions that are most likely to happen), we define the interaction score $s(u, v)$ by using the representations of users and items.
As we minimize the prediction error between $u$ and $v$ for each positive interaction $(u, v)$, their positive relationship is encoded into the $l_2$ distance between their representations (Equation~\eqref{eq:loss}).
In other words, a smaller value of $\mathcal{L}_{\theta, \xi} (u, v)$ indicates that the user-item pair $(u, v)$ is more likely to be interacted, which means the loss becomes inversely proportional to the interaction score.
To consider the symmetric relationship between $u$ and $v$, the interaction score is defined based on the cross-prediction task;
the prediction of $v$ by $u$, and the prediction of $u$ by $v$.\footnote{We empirically found that the normalized representations cannot take into account the popularity of users and items, thus simply use the output of the online encoder.}
\begin{equation}
\label{eq:score}
    s(u, v) = \predfn{\ouser}^\top \oitem +  \ouser^\top \predfn{\oitem}.
\end{equation}
For the computation of the interaction scores, we use only the representations obtained from the online encoder, with the target encoder discarded.
Since the online encoder and the target encoder finally converge to equilibrium by the slow-moving average, it is possible to effectively infer the interaction score only with the online encoder.
Considering the purpose of the target network, which generates the target for training the online network, it does make sense to leave the online encoder in the end.

Existing discriminative OCCF methods~\cite{rendle2009bpr, hsieh2017collaborative} have tried to optimize the latent space where the user-item interactions are directly encoded into their inner product (or Euclidean distance).
On the contrary, \proposed additionally uses the predictor to model their interaction, which results in the capability of encoding the high-level relationship between users and items into the representations.
In conclusion, with the help of the predictor, \proposed accurately computes the user-item interaction scores as well as optimizes the representation without explicitly using negative samples.

\begin{figure}[t]
    \centering
    \includegraphics[width=\linewidth]{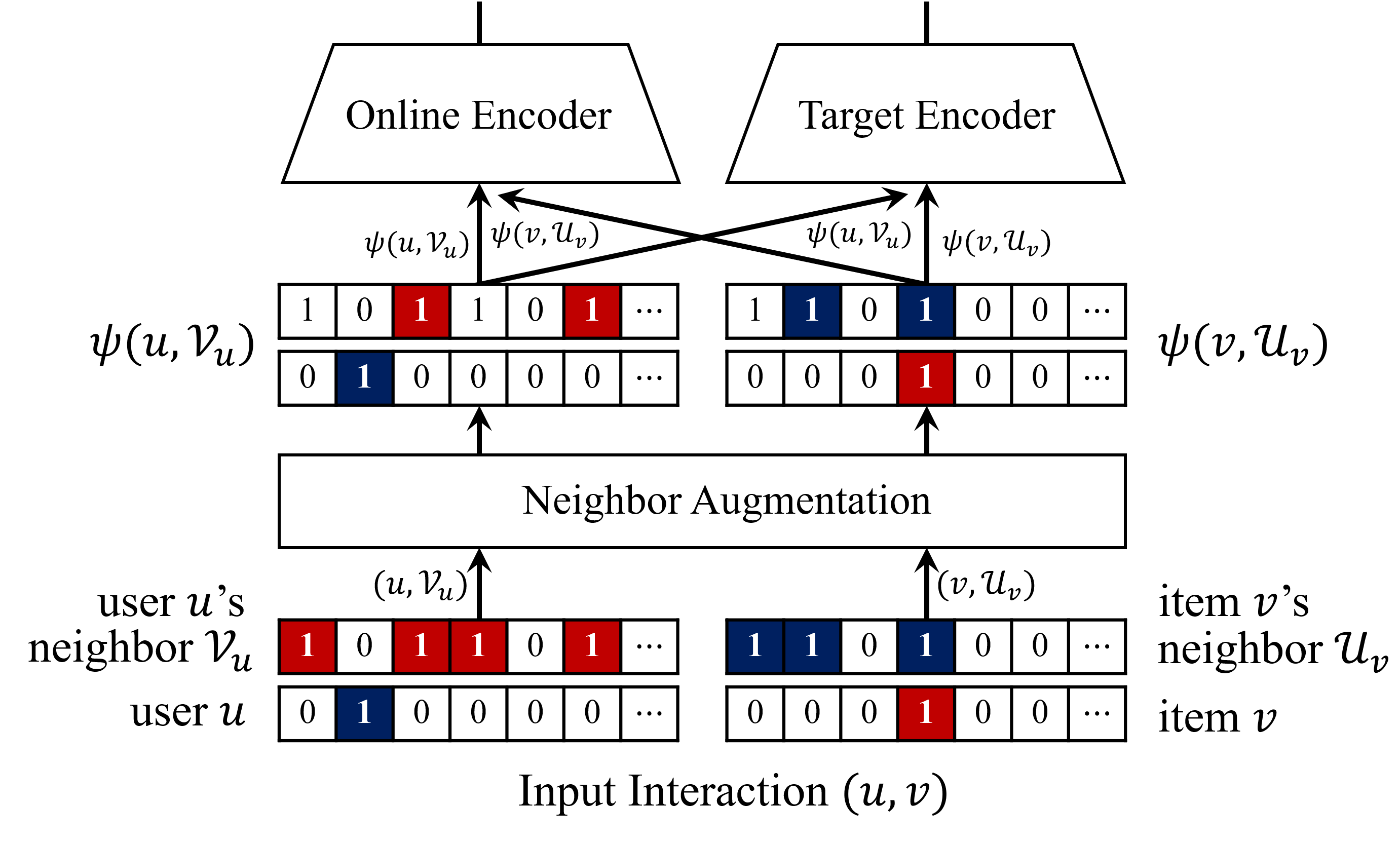}
    \caption{The stochastic data augmentation technique of \proposed based on the neighborhood information.}
    \label{fig:augmentation}
\end{figure}

\subsection{Neighbor-based Data Augmentation}
\label{subsec:nbaug}
The another available source for OCCF is the neighborhood information of users and items.
The neighbors of user $u$ and item $v$, denoted by $\userset$ and $\itemset$, refer to the set of the items interacted with $u$, and the users interacted with $v$, respectively.
From the perspective that user-item interactions can be considered as a bipartite graph between user nodes and item nodes, each node's neighbors (or its local graph structure) can be a good feature to encode the similarity among the nodes.
To take advantage of these neighbors as input features of users and items, we use a neighbor-based encoder~\cite{kim2019dual,wang2019neural,he2020lightgcn} which additionally takes a given set of users (or items) as its input.
Namely, this encoder is able to learn such set-featured inputs, represented as multi-hot vectors, capturing both the co-occurrence of users (or items) and their relationship. 
Adding the multi-hot inputs $\userset$ and $\itemset$ to the one-hot inputs $u$ and $v$ within our framework, the neighbor-based user/item representations, denoted by $\ouserset$ and $\oitemset$, can be effectively optimized and utilized, instead of $\ouser$ and $\oitem$.
In this case, the online encoder parameters related to user $u$ (or item $v$) are shared for computing $\ouserset$ and $\oitemset$, thus they are updated by two types of supervision (i.e., optimized not only as a target but also as one of the neighbors), which brings an effect of regularization.

For acquisition and exploitation of richer supervision, we extend our framework to consider much more user-item interactions that are augmented based on their neighborhood information in a self-supervised manner.
To this end, we introduce a new augmentation technique specifically designed for positive user-item interactions;
it does not statically increase the number of interactions as a pre-processing step, rather be \textit{stochastically} applied to each input interaction during the training.
This stochastic data augmentation allows the encoder to learn slightly perturbed interactions, referred to as \textit{augmented views} of an interaction.
By doing so, \proposed can effectively learn the representations even in the case that only a few positive user-item interactions are available for training (i.e., highly sparse dataset).
To this end, we first represent each user and item as the pair of its identity and neighbors: $(u, \userset)$ and $(v, \itemset)$.
Then, we apply the following augmentation function $\psi$ to the user and item before passing them to the neighbor encoder.
\begin{equation}
\begin{split}
   \small
    \psi(u, \userset) &= (u, \userset{}'), \text{ where } \userset{}' \sim \{\mathcal{S}|\mathcal{S}\subseteq \userset\}, \\
    \psi(v, \itemset) &= (v, \itemset{}'), \text{ where } \itemset{}' \sim \{\mathcal{S}|\mathcal{S}\subseteq \itemset\}. 
\end{split}
\end{equation}
This augmentation function chooses one of the subsets of the user's neighbors (i.e., $\userset{}'$) for an input user, and works in a similar way for an input item.
For each input interaction $(u, v)$, we can make a variety of interactions containing small perturbations $(\psi(u, \userset), \psi(v , \itemset))$, and they produce a similar effect to increasing the number of positive pairs from the data itself.

Similarly to Section~\ref{subsec:buir}, the online encoder is trained by minimizing $\mathcal{L}_{\theta,\xi}(\psi(u, \userset), \psi(v, \itemset))$, and the target encoder is slowly updated by the momentum mechanism.
After the optimization is finished, the interaction score is inferred by $\ouserset$ and $\oitemset$ (Equation~\eqref{eq:score}).
Figure~\ref{fig:augmentation} shows an example of our data augmentation which injects a certain level of perturbations to the neighbors.

\section{Experiments}
\label{sec:exp}
In this section, we describe the experimental results that support the superiority of our proposed framework.
We first present comparison results with other OCCF methods for top-$K$ recommendation (Section~\ref{subsec:exp_occf}), then validate the effectiveness of each component through an ablation study (Section~\ref{subsec:exp_ablation} and \ref{subsec:exp_augeffect}).
We also evaluate the quality of obtained representations for a downstream task (Section~\ref{subsec:exp_qualrep}) and finally provide the hyperparameter analysis (Section~\ref{subsec:exp_sensitivity}).


\subsection{Experimental Settings}
\label{subsec:exp_set}

\smallsection{Datasets}
In our experiments, we use three real-world datasets: \culike~\cite{wang2013collaborative}, \ciao~\cite{tang2012mtrust}, and \foursq~\cite{liu2017experimental}.
For preprocessing the datasets, we follow previous work~\cite{wang2019neural, he2017neural, rendle2009bpr, kang2020rrd} which provide the minimum count of user-item interactions for filtering long-tail users/items, considering the property of each dataset (e.g., the statistics or the domain where the implicit feedback is collected).\footnote{We remove users having fewer than 5 (\culike, \ciao) \& 20 interactions (\foursq), and remove items having fewer than 5 (\ciao) \& 10 interactions (\foursq).}
Table~\ref{tbl:datastats} summarizes the statistics of the datasets.


\begin{table}[t]
\caption{The statistics of the datasets.}
    \centering
    \begin{tabular}{r|rrr}
    \toprule
    Dataset & \multicolumn{1}{c}{\culike} & \multicolumn{1}{c}{\ciao} & \multicolumn{1}{c}{\foursq} \\\midrule
    \#Users & 5,219 & 7,265 & 19,465 \\
    \#Items & 25,181 & 11,211 & 28,593 \\
    \#Interactions & 125,580 & 149,141 & 1,115,108 \\
    Density & 0.096\% & 0.183\% & 0.200\% \\\bottomrule
    \end{tabular}
    \label{tbl:datastats}
\end{table}

\smallsection{Baselines}
We compare the performance of \proposed with that of baseline OCCF methods, including both discriminative and generative methods.
They are re-categorized as either 1) the methods using only the user-id/item-id or 2) the ones additionally using the neighborhood information.
Most of the methods in the first category directly optimize the embedding vectors of users and items.
\begin{itemize}
    \item \textbf{\bpr}~\cite{rendle2009bpr}:
    The Bayesian personalized ranking method for OCCF. It optimizes matrix factorization (MF) based on the pairwise ranking loss.
    \item \textbf{\neumf}~\cite{he2017neural}:
    The neural network-based method that uses the pointwise prediction loss. It combines MF and multi-layer perceptron (MLP) to model the user-item interaction.
    \item \textbf{\cml}~\cite{hsieh2017collaborative}:
    A metric learning approach to the OCCF problem. It optimizes the Euclidean distance between a user and an item based on the pairwise hinge loss. 
    \item \textbf{\sml}~\cite{li2020symmetric}:
    The state-of-the-art OCCF method based on metric learning.
    For symmetrization, it considers the Euclidean distance among items as well as between a user and an item.
\end{itemize}

Next, the neighbor-based OCCF methods exploit the neighborhood information of users and items to compute the representations.
\begin{itemize}
    
    \item \textbf{\ngcf}~\cite{wang2019neural}:
    A neighbor-based method which encodes a user's (and item's) neighbors by using graph convolutional networks (GCN). 
    It can consider multi-hop neighbors as well based on a stack of GCN layers.
    
    \item \textbf{\lgcn}~\cite{he2020lightgcn}:
    The state-of-the-art method that further tailors the GCN-based user (and item) encoder for the OCCF task.
    It simplifies the GCN by using the light graph convolution.  
    
    \item \textbf{\vae}~\cite{liang2018variational}: 
    The OCCF method based on a variational autoencoder that reconstructs partially-observed user vectors.
    It enforces the latent distribution to approximate the prior, assumed to be the normal distribution.
    
    \item \textbf{\cfgan}~\cite{chae2018cfgan}: 
    The state-of-the-art GAN-based OCCF method.
    The discriminator is trained to distinguish between input (real) user vectors and generated (fake) ones, while the generator is optimized to deceive the discriminator.
\end{itemize}
Among them, \ngcf and \lgcn are the discriminative methods that optimize their model by using the pairwise loss based on the \bpr framework.
On the contrary, \vae and \cfgan are the generative methods that focus on learning the latent distribution of users, represented by binary vectors indicating their interacted items.

We build two variants of \proposed using different encoder networks.
\begin{itemize}
    \item \textbf{\proposedid}: 
    The \proposed framework using a single embedding layer as its encoder. It simply takes the user/item vectors from the embedding matrix (Section~\ref{subsec:buir}).
    
    \item \textbf{\proposednb}: 
    The \proposed framework based on the \lgcn encoder.
    It computes the user/item representations by using the lightweight GCN~\cite{he2020lightgcn} that adopts the proposed neighbor augmentation technique (Section~\ref{subsec:nbaug}).
\end{itemize}
Note that any types of user/item encoder networks, which are originally optimized in a discriminative framework (e.g., BPR), can be easily embedded into our framework.

\smallsection{Evaluation Protocols}
For each dataset, we randomly split each user’s interaction history into training/validation/test sets, with various split ratios.
In detail, to verify the effectiveness of \proposed with varying levels of data sparsity, we build three training sets that include a certain proportion of interactions for each user, i.e., $\beta \in \{ 10\%, 20\%, 50\%\}$,\footnote{This setting (high sparsity) is more difficult and practical than the traditional setting.} then equally divide the rest into the validation set and the test set.
We report the average value of five independent runs, each of which uses different random seeds for the split.

As we focus on the top-$K$ recommendation task for implicit feedback, we evaluate the performance of each method by using two widely-used ranking metrics~\cite{chae2018cfgan, liang2018variational, li2020symmetric}: Precision (P@$K$) and Normalized Discounted Cumulative Gain (N@$K$).\footnote{As pointed out in~\cite{krichene2020sampled}, a sampled metric where only a smaller set of random items and the relevant items are ranked (e.g., leave-one-out evaluation protocol~\cite{he2017neural}) cannot correctly indicate the true performance of recommender systems. For this reason, we instead consider the ranked list of all the items with no interaction.}
P@$K$ measures how many test items are included in the list of top-$K$ items and N@$K$ assigns higher scores on the upper-ranked test items. 

\smallsection{Implementation Details}
We implement the proposed framework and all the baselines by using PyTorch, and use the Adam optimizer to train them.
For \proposed, we fix the momentum coefficient $\tau$ to 0.995, and adopt a single linear layer for the predictor $q_\theta$.\footnote{We empirically found that these hyperparameters hardly affect the final performance of \proposed, and the sensitivity analysis on the parameters is provided in Section~\ref{subsec:exp_sensitivity}.}
The augmentation function $\psi$ simply uses a uniform distribution for drawing a drop probability $p\sim\mathcal{U}(0, 1)$, where each user's (item's) neighbor is independently deleted with the probability $p$.

For each dataset and baseline, we tune the hyperparameters using a grid search, which finds their optimal values that achieve the best performance on the validation set:
the dimension size of representations $D\in\{50, 100, 150, 200, 250\}$, the weight decay (i.e., coefficient for $L_2$ regularization) $\lambda\in\{10^{-1}, 10^{-2}, 10^{-3}, 10^{-4}, 10^{-5}\}$, the initial learning rate $\eta\in\{10^{-1}, 10^{-2}, 10^{-3} 10^{-4}\}$, and the number of negative pairs for each positive pair (particularly for discriminative baselines) $n\in\{1, 2, 5, 10, 20\}$.
In case of baseline-specific hyperparameters, we tune them in the ranges suggested by their original papers.
We set the maximum number of epochs to 500 and adopt the early stopping strategy;
it terminates when P@10 on the validation set does not increase for 50 successive epochs.

\subsection{Comparison with OCCF Methods}
\label{subsec:exp_occf}
\renewcommand{\arraystretch}{0.8}
\begin{table*}[t]
\caption{The recommendation performances of a wide range of OCCF methods, varying the sparsity of the datasets. \textit{Improv}\textsubscript{id} and \textit{Improv}\textsubscript{nb} respectively denote the improvement of \proposed over the best id/neighbor-based baseline. The superscripts *, **, and *** indicate $p\leq 0.05$, $p\leq 0.005$, and $p\leq 0.0005$ for the paired t-test of \proposednb vs. the best baseline on P@10. }
\label{tbl:mainresults}
\resizebox{0.99\linewidth}{!}{%
\begin{tabular}{cPcPPPPPPPPPPPP}
\toprule
\multicolumn{3}{c}{Setting} & \multicolumn{6}{c}{User/Item ID} & \multicolumn{6}{c}{User/Item ID + Neighbor} \\
\cmidrule(lr){1-3}\cmidrule(lr){4-9}\cmidrule(lr){10-15}
Data & $\beta$ & Metric & \bpr & \neumf & \cml & \sml & \proposedid & {\small \textit{Improv}\textsubscript{id}} & \ngcf & \lgcn & {\small \vae} & {\small \cfgan} & \proposednb & {\small \textit{Improv}\textsubscript{nb}} \\ 
\midrule
\multirow{20}{*}{\rotatebox{90}{\culike}} & \multirow{6.5}{*}{\textsuperscript{\ \ }10\%\textsuperscript{***}}  
                                                    & P@10 & 0.0369 & 0.0350 & 0.0327 & 0.0279 & \textbf{0.0542} & 46.88\% & 0.0387 & 0.0518 & 0.0330 & 0.0437 & \textbf{0.0637} & 22.80\% \\
                            &                       & P@20 & 0.0484 & 0.0474 & 0.0451 & 0.0409 & \textbf{0.0708} & 46.20\% & 0.0506 & 0.0676 & 0.0444 & 0.0589 & \textbf{0.0814} & 20.38\% \\
                            &                       & P@50 & 0.0729 & 0.0785 & 0.0790 & 0.0685 & \textbf{0.1050} & 32.93\% & 0.0762 & 0.1010 & 0.0740 & 0.0968 & \textbf{0.1202} & 19.08\% \\
                            \cmidrule(lr){3-3}\cmidrule(lr){4-9}\cmidrule(lr){10-15}
                            &                       & N@10 & 0.0310 & 0.0311 & 0.0272 & 0.0222 & \textbf{0.0480} & 54.60\% & 0.0337 & 0.0465 & 0.0289 & 0.0382 & \textbf{0.0568} & 22.11\% \\
                            &                       & N@20 & 0.0351 & 0.0349 & 0.0316 & 0.0266 & \textbf{0.0533} & 51.88\% & 0.0376 & 0.0516 & 0.0327 & 0.0433 & \textbf{0.0623} & 20.78\% \\
                            &                       & N@50 & 0.0429 & 0.0441 & 0.0421 & 0.0350 & \textbf{0.0636} & 44.18\% & 0.0456 & 0.0619 & 0.0417 & 0.0548 & \textbf{0.0742} & 19.91\% \\
                            \cmidrule(lr){2-3}\cmidrule(lr){4-9}\cmidrule(lr){10-15}
                            & \multirow{6.5}{*}{\textsuperscript{\ \ }20\%\textsuperscript{***}} 
                                                    & P@10 & 0.0634 & 0.0422 & 0.0696 & 0.0515 & \textbf{0.0903} & 29.78\% & 0.0684 & 0.0835 & 0.0433 & 0.0730 & \textbf{0.0956} & 14.48\% \\
                            &                       & P@20 & 0.0862 & 0.0565 & 0.0964 & 0.0717 & \textbf{0.1210} & 25.41\% & 0.0915 & 0.1097 & 0.0601 & 0.0979 & \textbf{0.1243} & 13.37\% \\
                            &                       & P@50 & 0.1298 & 0.0847 & 0.1506 & 0.1145 & \textbf{0.1775} & 17.83\% & 0.1356 & 0.1607 & 0.0973 & 0.1237 & \textbf{0.1807} & 12.45\% \\
                            \cmidrule(lr){3-3}\cmidrule(lr){4-9}\cmidrule(lr){10-15}
                            &                       & N@10 & 0.0510 & 0.0358 & 0.0576 & 0.0424 & \textbf{0.0795} & 37.93\% & 0.0580 & 0.0727 & 0.0377 & 0.0566 & \textbf{0.0831} & 14.22\% \\
                            &                       & N@20 & 0.0591 & 0.0407 & 0.0668 & 0.0494 & \textbf{0.0880} & 31.74\% & 0.0657 & 0.0812 & 0.0435 & 0.0631 & \textbf{0.0912} & 12.40\% \\
                            &                       & N@50 & 0.0726 & 0.0493 & 0.0833 & 0.0627 & \textbf{0.1050} & 25.96\% & 0.0793 & 0.0966 & 0.0548 & 0.0754 & \textbf{0.1071} & 10.82\% \\
                            \cmidrule(lr){2-3}\cmidrule(lr){4-9}\cmidrule(lr){10-15}
                            & \multirow{6.5}{*}{\textsuperscript{\ \ }50\%\textsuperscript{**\ \ }} 
                                                    & P@10 & 0.1229 & 0.1138 & 0.1310 & 0.1195 & \textbf{0.1555} & 18.73\% & 0.1470 & 0.1561 & 0.1116 & 0.1389 & \textbf{0.1624} & \ \ 4.05\%  \\
                            &                       & P@20 & 0.1719 & 0.1512 & 0.1845 & 0.1690 & \textbf{0.2065} & 11.91\% & 0.1978 & 0.2110 & 0.1513 & 0.1863 & \textbf{0.2170} & \ \ 2.84\%  \\
                            &                       & P@50 & 0.2566 & 0.2162 & 0.2794 & 0.2545 & \textbf{0.2993} & \ \ 7.12\%  & 0.2862 & 0.3056 & 0.2243 & 0.2677 & \textbf{0.3120} & \ \ 2.10\%  \\
                            \cmidrule(lr){3-3}\cmidrule(lr){4-9}\cmidrule(lr){10-15}
                            &                       & N@10 & 0.0891 & 0.0877 & 0.0950 & 0.0899 & \textbf{0.1189} & 25.23\% & 0.1122 & 0.1189 & 0.0843 & 0.1052 & \textbf{0.1240} & \ \ 4.29\%  \\
                            &                       & N@20 & 0.1046 & 0.0994 & 0.1121 & 0.1055 & \textbf{0.1348} & 20.25\% & 0.1283 & 0.1360 & 0.0968 & 0.1201 & \textbf{0.1405} & \ \ 3.29\%  \\
                            &                       & N@50 & 0.1276 & 0.1174 & 0.1379 & 0.1287 & \textbf{0.1600} & 16.06\% & 0.1525 & 0.1617 & 0.1169 & 0.1425 & \textbf{0.1656} & \ \ 2.40\%  \\
\midrule
\multirow{20}{*}{\rotatebox{90}{\ciao}} & \multirow{6.5}{*}{\textsuperscript{\ \ }10\%\textsuperscript{***}} 
                                               & P@10 & 0.0289 & 0.0302 & 0.0422 & 0.0461 & \textbf{0.0598} & 29.67\% & 0.0336 & 0.0582 & 0.0434 & 0.0521 & \textbf{0.0664} & 14.05\% \\
                       &                       & P@20 & 0.0346 & 0.0404 & 0.0603 & 0.0697 & \textbf{0.0787} & 12.97\% & 0.0430 & 0.0748 & 0.0573 & 0.0679 & \textbf{0.0831} & 11.07\% \\
                       &                       & P@50 & 0.0508 & 0.0627 & 0.1021 & 0.1043 & \textbf{0.1123} & \ \ 7.59\%  & 0.0669 & 0.1095 & 0.0843 & 0.0972 & \textbf{0.1177} & \ \ 7.47\%  \\
                       \cmidrule(lr){3-3}\cmidrule(lr){4-9}\cmidrule(lr){10-15}
                       &                       & N@10 & 0.0278 & 0.0269 & 0.0369 & 0.0418 & \textbf{0.0535} & 27.96\% & 0.0313 & 0.0557 & 0.0391 & 0.0443 & \textbf{0.0628} & 12.75\% \\
                       &                       & N@20 & 0.0289 & 0.0301 & 0.0433 & 0.0506 & \textbf{0.0588} & 16.16\% & 0.0339 & 0.0597 & 0.0434 & 0.0479 & \textbf{0.0675} & 12.99\% \\
                       &                       & N@50 & 0.0337 & 0.0371 & 0.0566 & 0.0643 & \textbf{0.0695} & \ \ 8.02\%  & 0.0415 & 0.0705 & 0.0519 & 0.0572 & \textbf{0.0784} & 11.17\% \\
                       \cmidrule(lr){2-3}\cmidrule(lr){4-9}\cmidrule(lr){10-15}
                       & \multirow{6.5}{*}{\textsuperscript{\ \ }20\%\textsuperscript{**\ \ }} 
                                               & P@10 & 0.0478 & 0.0361 & 0.0505 & 0.0517 & \textbf{0.0608} & 17.62\% & 0.0440 & 0.0662 & 0.0501 & 0.0535 & \textbf{0.0724} & \ \ 9.37\%  \\
                       &                       & P@20 & 0.0623 & 0.0469 & 0.0728 & 0.0725 & \textbf{0.0817} & 12.19\% & 0.0580 & 0.0849 & 0.0656 & 0.0705 & \textbf{0.0911} & \ \ 7.30\%  \\
                       &                       & P@50 & 0.0940 & 0.0711 & 0.1126 & 0.1089 & \textbf{0.1210} & \ \ 7.44\%  & 0.0886 & 0.1247 & 0.0993 & 0.1090 & \textbf{0.1322} & \ \ 5.98\%  \\
                       \cmidrule(lr){3-3}\cmidrule(lr){4-9}\cmidrule(lr){10-15}
                       &                       & N@10 & 0.0436 & 0.0322 & 0.0432 & 0.0419 & \textbf{0.0539} & 23.68\% & 0.0403 & 0.0606 & 0.0445 & 0.0480 & \textbf{0.0670} & 10.52\% \\
                       &                       & N@20 & 0.0481 & 0.0354 & 0.0514 & 0.0492 & \textbf{0.0598} & 16.17\% & 0.0447 & 0.0659 & 0.0491 & 0.0525 & \textbf{0.0725} & 10.08\% \\
                       &                       & N@50 & 0.0578 & 0.0429 & 0.0666 & 0.0605 & \textbf{0.0726} & \ \ 8.98\%  & 0.0546 & 0.0786 & 0.0596 & 0.0619 & \textbf{0.0854} & \ \ 8.71\%  \\
                       \cmidrule(lr){2-3}\cmidrule(lr){4-9}\cmidrule(lr){10-15}
                       & \multirow{6.5}{*}{\textsuperscript{\ \ }50\%\textsuperscript{**\ \ }} 
                                               & P@10 & 0.0679 & 0.0426 & 0.0533 & 0.0639 & \textbf{0.0786} & 15.83\% & 0.0576 & 0.0746 & 0.0481 & 0.0705 & \textbf{0.0812} & \ \ 8.79\%  \\
                       &                       & P@20 & 0.0909 & 0.0602 & 0.0875 & 0.0873 & \textbf{0.1005} & 10.53\% & 0.0831 & 0.1029 & 0.0676 & 0.0982 & \textbf{0.1092} & \ \ 6.10\%  \\
                       &                       & P@50 & 0.1370 & 0.0924 & 0.1579 & 0.1279 & \textbf{0.1642} & \ \ 3.99\%  & 0.1302 & 0.1578 & 0.1000 & 0.1502 & \textbf{0.1673} & \ \ 6.02\%  \\
                       \cmidrule(lr){3-3}\cmidrule(lr){4-9}\cmidrule(lr){10-15}
                       &                       & N@10 & 0.0563 & 0.0337 & 0.0394 & 0.0530 & \textbf{0.0641} & 13.86\% & 0.0463 & 0.0612 & 0.0391 & 0.0570 & \textbf{0.0679} & 10.98\% \\
                       &                       & N@20 & 0.0633 & 0.0393 & 0.0516 & 0.0601 & \textbf{0.0706} & 11.43\% & 0.0546 & 0.0702 & 0.0456 & 0.0659 & \textbf{0.0766} & \ \ 9.18\%  \\
                       &                       & N@50 & 0.0767 & 0.0485 & 0.0725 & 0.0718 & \textbf{0.0824} & \ \ 7.54\%  & 0.0685 & 0.0862 & 0.0551 & 0.0810 & \textbf{0.0932} & \ \ 8.17\%  \\
\midrule
\multirow{20}{*}{\rotatebox{90}{\foursq}} & \multirow{6.5}{*}{\textsuperscript{\ \ }10\%\textsuperscript{***}} 
                                                     & P@10 & 0.0561 & 0.0441 & 0.0451 & 0.0317 & \textbf{0.0890} & 58.70\% & 0.0658 & 0.0926 & 0.0451 & 0.0519 & \textbf{0.0999} & \ \ 7.84\% \\
                             &                       & P@20 & 0.0696 & 0.0528 & 0.0623 & 0.0421 & \textbf{0.1020} & 46.65\% & 0.0793 & 0.1054 & 0.0534 & 0.0612 & \textbf{0.1127} & \ \ 6.97\% \\
                             &                       & P@50 & 0.1173 & 0.0758 & 0.1180 & 0.0698 & \textbf{0.1573} & 33.32\% & 0.1305 & 0.1613 & 0.0746 & 0.0981 & \textbf{0.1695} & \ \ 5.06\% \\
                             \cmidrule(lr){3-3}\cmidrule(lr){4-9}\cmidrule(lr){10-15}
                             &                       & N@10 & 0.0619 & 0.0476 & 0.0451 & 0.0311 & \textbf{0.1023} & 65.27\% & 0.0732 & 0.1080 & 0.0483 & 0.0579 & \textbf{0.1161} & \ \ 7.44\% \\
                             &                       & N@20 & 0.0665 & 0.0503 & 0.0531 & 0.0360 & \textbf{0.1045} & 57.27\% & 0.0771 & 0.1104 & 0.0508 & 0.0610 & \textbf{0.1171} & \ \ 6.05\% \\
                             &                       & N@50 & 0.0865 & 0.0600 & 0.0769 & 0.0480 & \textbf{0.1277} & 47.52\% & 0.0985 & 0.1336 & 0.0598 & 0.0770 & \textbf{0.1409} & \ \ 5.46\% \\
                             \cmidrule(lr){2-3}\cmidrule(lr){4-9}\cmidrule(lr){10-15}
                             & \multirow{6.5}{*}{\textsuperscript{\ \ }20\%\textsuperscript{***}} 
                                                     & P@10 & 0.0752 & 0.0489 & 0.0754 & 0.0820 & \textbf{0.1099} & 34.11\% & 0.0872 & 0.1063 & 0.0658 & 0.0856 & \textbf{0.1142} & \ \ 7.43\% \\
                             &                       & P@20 & 0.0941 & 0.0633 & 0.0988 & 0.0985 & \textbf{0.1281} & 29.75\% & 0.1065 & 0.1239 & 0.0779 & 0.1035 & \textbf{0.1323} & \ \ 6.82\% \\
                             &                       & P@50 & 0.1573 & 0.1099 & 0.1714 & 0.1515 & \textbf{0.1997} & 16.47\% & 0.1733 & 0.1905 & 0.1220 & 0.1643 & \textbf{0.2043} & \ \ 7.24\% \\
                             \cmidrule(lr){3-3}\cmidrule(lr){4-9}\cmidrule(lr){10-15}
                             &                       & N@10 & 0.0829 & 0.0532 & 0.0785 & 0.0833 & \textbf{0.1273} & 52.77\% & 0.0973 & 0.1247 & 0.0725 & 0.1003 & \textbf{0.1328} & \ \ 6.47\% \\
                             &                       & N@20 & 0.0898 & 0.0588 & 0.0888 & 0.0904 & \textbf{0.1312} & 45.07\% & 0.1036 & 0.1281 & 0.0760 & 0.1060 & \textbf{0.1363} & \ \ 6.42\% \\
                             &                       & N@50 & 0.1161 & 0.0782 & 0.1195 & 0.1129 & \textbf{0.1607} & 34.48\% & 0.1313 & 0.1564 & 0.0949 & 0.1321 & \textbf{0.1660} & \ \ 6.17\% \\
                             \cmidrule(lr){2-3}\cmidrule(lr){4-9}\cmidrule(lr){10-15}
                             & \multirow{6.5}{*}{\textsuperscript{\ \ }50\%\textsuperscript{***}} 
                                                     & P@10 & 0.0894 & 0.0900 & 0.0843 & 0.1005 & \textbf{0.1125} & 12.00\% & 0.1064 & 0.1123 & 0.0838 & 0.0965 & \textbf{0.1204} & \ \ 7.25\% \\
                             &                       & P@20 & 0.1249 & 0.1237 & 0.1202 & 0.1404 & \textbf{0.1546} & 10.16\% & 0.1469 & 0.1509 & 0.1226 & 0.1325 & \textbf{0.1595} & \ \ 5.71\% \\
                             &                       & P@50 & 0.2059 & 0.2024 & 0.2025 & 0.2273 & \textbf{0.2502} & 10.07\% & 0.2373 & 0.2386 & 0.2086 & 0.2162 & \textbf{0.2531} & \ \ 6.09\% \\
                             \cmidrule(lr){3-3}\cmidrule(lr){4-9}\cmidrule(lr){10-15}
                             &                       & N@10 & 0.0898 & 0.0919 & 0.0830 & 0.1002 & \textbf{0.1152} & 14.96\% & 0.1071 & 0.1153 & 0.0796 & 0.1007 & \textbf{0.1211} & \ \ 5.01\% \\
                             &                       & N@20 & 0.1046 & 0.1058 & 0.0979 & 0.1166 & \textbf{0.1324} & 13.55\% & 0.1241 & 0.1307 & 0.0962 & 0.1152 & \textbf{0.1361} & \ \ 4.12\% \\
                             &                       & N@50 & 0.1341 & 0.1344 & 0.1279 & 0.1484 & \textbf{0.1672} & 12.70\% & 0.1571 & 0.1657 & 0.1274 & 0.1456 & \textbf{0.1709} & \ \ 3.14\% \\
\bottomrule
\end{tabular}
}
\end{table*}
We first measure the top-$K$ recommendation performance of \proposed and the baseline methods.
Table~\ref{tbl:mainresults} presents the comparison results on three different sparsity levels of datasets.
In summary, \proposed achieves the best performance among all the baselines, and especially shows the significant improvements in highly sparse datasets.
We analyze the results from various perspectives.

\subsubsection{Effectiveness of \proposedid}
\label{subsubsec:proposedid}
For all the datasets, \proposedid shows the substantially higher performance than the discriminative methods taking only user-id/item-id (i.e., \bpr, \neumf, \cml, and \sml).
In particular, the sparser the training set becomes, the larger the performance improvement of \proposedid is achieved over the best baseline (denoted by \textit{Improv}\textsubscript{id}).
It is obvious that \proposedid is more robust to the extreme sparsity compared to the other baselines that are more likely to explicitly use ``positive but unobserved'' interactions as negative interactions when positive user-item interactions are more rarely observed.
\proposedid is not affected by such inconsistent supervision from uncertain negative interactions because it directly optimizes the representations of users and items by using only positive interactions.

Furthermore, in terms of the number of retrieved items (denoted by $K$), \proposed shows much larger performance improvements for P@10 and N@10 compared to P@50 and N@50, respectively.
In other words, \proposed performs much better at predicting the top-ranked items than the other baselines, which makes it practically advantageous for real-world recommender systems that aim to accurately provide the most preferred items to their users.


\begin{figure}[t]
	\centering
    \includegraphics[width=\linewidth]{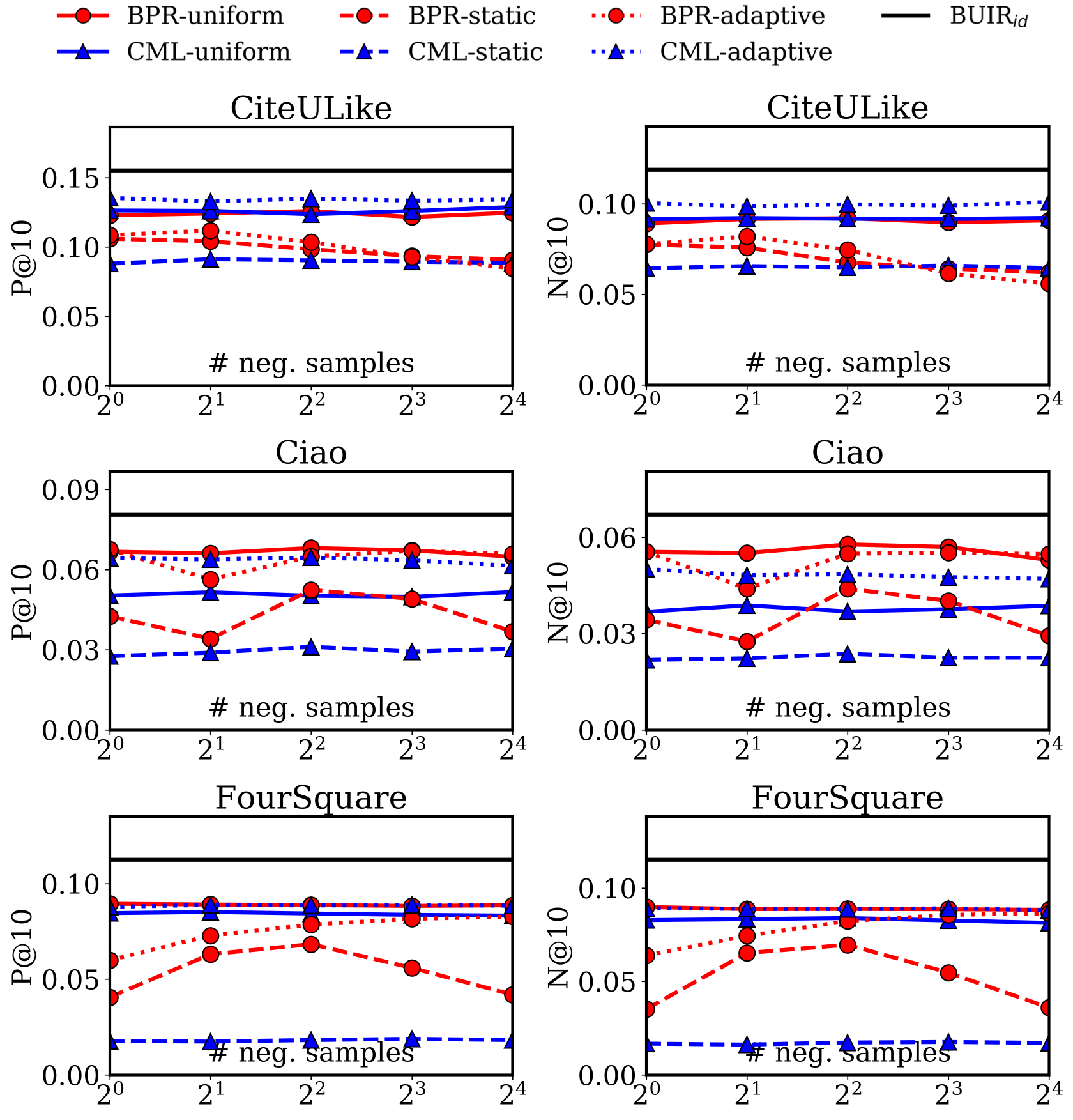}
	\caption{Comparison with discriminative methods (\bpr and \cml) using various negative sampling strategies.} 
	\label{fig:sampling_effect}
\end{figure}

\subsubsection{Effectiveness of \proposednb}
\label{subsubsec:proposednb}
We also observe that \proposednb significantly outperforms all the other neighbor-based competitors, including discriminative (i.e., \ngcf and \lgcn) and generative methods (i.e., \vae and \cfgan).
Similar to Section~\ref{subsubsec:proposedid}, there exist a consistent trend on its performance gain (denoted by \textit{Improv}\textsubscript{nb}), which becomes more significant as fewer interactions are given for training.
Specifically, the neighbor-based baselines improve the recommendation performance over the methods not using the neighborhood information, as they are able to cope with the high sparsity to some degree by leveraging the neighbors of users and items.
Nevertheless, most of them, except for \lgcn, perform worse than even \proposedid; 
this strongly indicates that their imperfect assumption on negative interactions severely limits the capability of capturing users' preference on items even though they utilize rich information sources as well as employ advanced neural architectures.
In short, for the OCCF problem where only a small number of positive interactions are given, our \proposed framework is effective regardless of the information sources used for training, in that any assumption on negative interactions is not required. 

In addition, the critical drawback of the generative methods is the difficulty of stable optimization.
For example, \vae should carefully treat the annealing technique for minimizing Kullback-Leibler (KL) divergence, 
and \cfgan needs to balance the adversarial updates between the discriminator and generator for their convergence to the equilibrium.
In contrast, \proposed can easily train the encoder without any advanced techniques for stable optimization, which makes our framework much practical.

\begin{table}[t]
\caption{Performances of \proposed that ablates each component.}
\label{tbl:ablation}
\centering
\resizebox{0.99\linewidth}{!}{%
    \begin{tabular}{c|TSSS|c}
    \toprule
    Method & \multicolumn{4}{c|}{\footnotesize Framework\ \ Predictor\ \ Neighbor\ \ Augment} & P@10 \\\midrule
    {\bpr} & \bpr & & & & 0.1229 $\pm$ 0.0035 \\
    & \bpr & \checkmark & & & 0.0752 $\pm$ 0.0027 \\
    {\lgcn} & \bpr & & \checkmark  & & 0.1561 $\pm$ 0.0038 \\\midrule
    {\proposedid} & \proposed & \checkmark & & & 0.1555 $\pm$ 0.0029 \\
    & \proposed & \checkmark & \checkmark  & & 0.1592 $\pm$ 0.0028 \\
    {\proposednb} & \proposed & \checkmark & \checkmark & \checkmark & \textbf{0.1624} $\pm$ 0.0032 \\\bottomrule
    \end{tabular}
}
\end{table}

\subsubsection{Comparison of different negative sampling strategies}
\label{subsubsec:exp_negsampling}

To examine how much the choice of a negative sampling strategy affects the recommendation performance, we measure P@10 and N@10 of two discriminative methods (i.e., \bpr and \cml) that adopt different strategies.
We vary the number of negative pairs (sampled for each positive pair) in the range of $\{2^0, 2^1, 2^2, 2^3, 2^4\}$, and consider three different distributions for negative sampling~\cite{rendle2014improving}:
1) \textit{uniform} sampling, 2) \textit{static-and-global} sampling which draws a pair based on the item popularity, and 3) \textit{adaptive-and-contextual} sampling that uses the probability proportional to the interaction score.

In Figure~\ref{fig:sampling_effect}, we observe that the performance of the discriminative methods largely depends on the sampling strategy, whereas \proposedid consistently performs the best.
To be specific, the sampling strategies show different tendencies or have different optimal hyperparameter values, depending on each dataset or each method.
For instance, \cml achieves marginal performance gains from the adaptive-and-contextual sampling compared to the uniform sampling, whereas \bpr does not take any benefits from it.
This is because \cml optimizes its model by the hinge loss, which cannot produce the gradients to update the model parameters for too easily-distinguishable negative pairs.
In this case, the adaptive-and-contextual sampling strategy can effectively select the hard-negative pairs for training, which accelerates the convergence and its final performance.
We remark that this kind of sampling techniques can improve the performance of the discriminative methods to some extent, but the sampling operation requires a high computational cost itself as well as the process of hyperparameter tuning for each dataset (and method) takes huge efforts.
On the contrary, as \proposedid does not rely on negative sampling, it always shows the greater performance (plotted as a solid black line) compared to any of the discriminative methods using various sampling techniques.
This result clearly validates the superiority of \proposed in that it is not affected by the choice of the negative sampling strategy any longer.

\subsection{Ablation Study}
\label{subsec:exp_ablation}
To validate the effectiveness of each component in our framework, we measure the performance of the methods that ablate the following components:
1) modeling the interaction score based on the predictor (i.e., cross-prediction score defined in Equation~\eqref{eq:score}), 2) the neighbor-based encoder that is able to capture the user's (item's) neighborhood information, and 3) the stochastic neighbor augmentation that produces various views of an input interaction.
In Table~\ref{tbl:ablation}, we report P@10 on the \culike dataset ($\beta$=50\%).

First of all, the \bpr framework that optimizes the cross-prediction score, $q\left(f(u)\right)^\top f(v) + f(u)^\top q\left(f(v)\right)$, is not as effective as ours;
it is even worse compared to the conventional \bpr, which optimizes the inner-product score $f(u)^\top f(v)$.
This implies that the performance improvement of \proposed is mainly caused by our learning framework rather than its score modeling based on the predictor.
In addition, even without the stochastic augmentation, the neighbor-based encoder (i.e., \lgcn) based on the \proposed framework beats \lgcn based on the BPR framework, which demonstrates that \proposed successfully addresses the issue of incorrect negative sampling.
Lastly, our framework with the stochastic neighbor augmentation further improves the performance by taking benefits from various views of the positive user-item interactions for the optimization.

\begin{figure}[t]
	\centering
    \includegraphics[width=\linewidth]{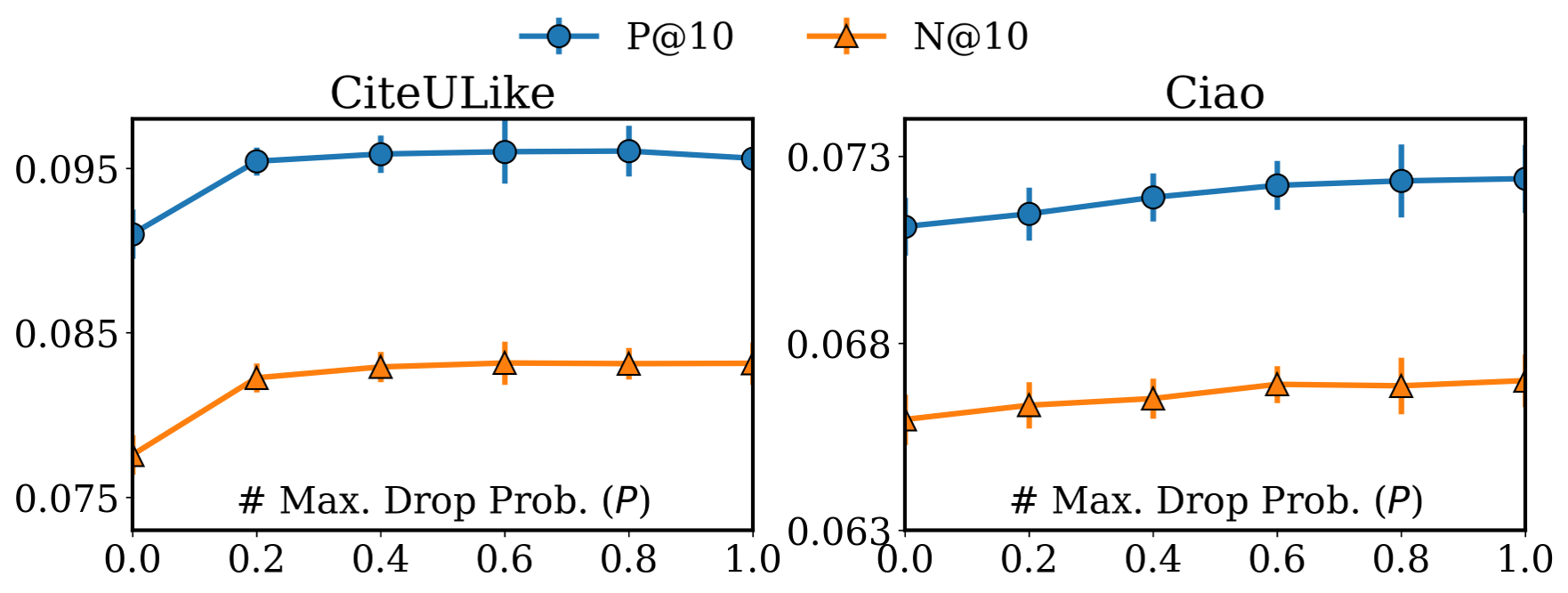}
	\caption{Performance changes of \proposednb with respect to the maximum drop probability for the augmentation.} 
	\label{fig:dropprob_effect}
\end{figure}

\subsection{Effect of Neighbor Augmentation}
\label{subsec:exp_augeffect}
For an in-depth analysis on the effect of our stochastic data augmentation function $\psi$, we measure the performance of \proposednb on the \culike and \ciao datasets ($\beta$=20\%), with various magnitudes of the perturbation added to the neighbors of users and items.
We modify the augmentation function to randomly select the drop probability from a predefined interval, i.e., $p\sim\mathcal{U}(0, P)$ where $P$ is the maximum drop probability, then increase $P$ from 0.0 to 1.0.

In Figure~\ref{fig:dropprob_effect}, our stochastic data augmentation (i.e., $P>0$) brings a significant improvement compared to the case of using the fixed neighborhood information (i.e., $P=0$) as encoder inputs.
This result shows that the augmented views of positive interactions encourage \proposed to effectively learn users' preference on items even in much sparse dataset.
Interestingly, in case of the \ciao dataset which is less sparse than \culike, the benefit of our augmentation linearly increases with the maximum drop probability.
This is because there is room for producing more various views (i.e., larger perturbation) based on a relatively more number of neighbors, and it eventually helps to boost the recommendation performance.
To sum up, our framework that adopts the neighbor augmentation function successfully relieves the data sparsity issue of the OCCF problem, by leveraging the augmented views of few positive interactions.

\subsection{Evaluation on Representation Quality}
\label{subsec:exp_qualrep}
To evaluate the quality of the obtained representations, we compare the performance for a downstream task by using the representations optimized by \proposed and the other baselines.\footnote{In this comparison, we exclude the generative OCCF methods as our baselines, because they do not explicitly output the item representations.}
We consider an item classification task to evaluate how well each method encodes the items' characteristics or latent semantics into the representations.
We choose two datasets that offer the side information on items, which are \ciao and \foursq.
\ciao provides the 28-category label of each item (i.e., the products), and \foursq contains the GPS coordinates for each item (i.e., point-of-interest).
In case of \foursq, we first perform $k$-means clustering on the coordinates with $k$=100, and use the clustering results as the class labels.
We train a linear and non-linear classifier (i.e., a single-layer perceptron and three-layer perceptron, respectively) to predict the class label of each item by using the fixed item representations as the input.
Finally, we perform 10-fold cross-validation and report the average result and standard deviation.

\begin{figure}[t]
	\centering
    \includegraphics[width=\linewidth]{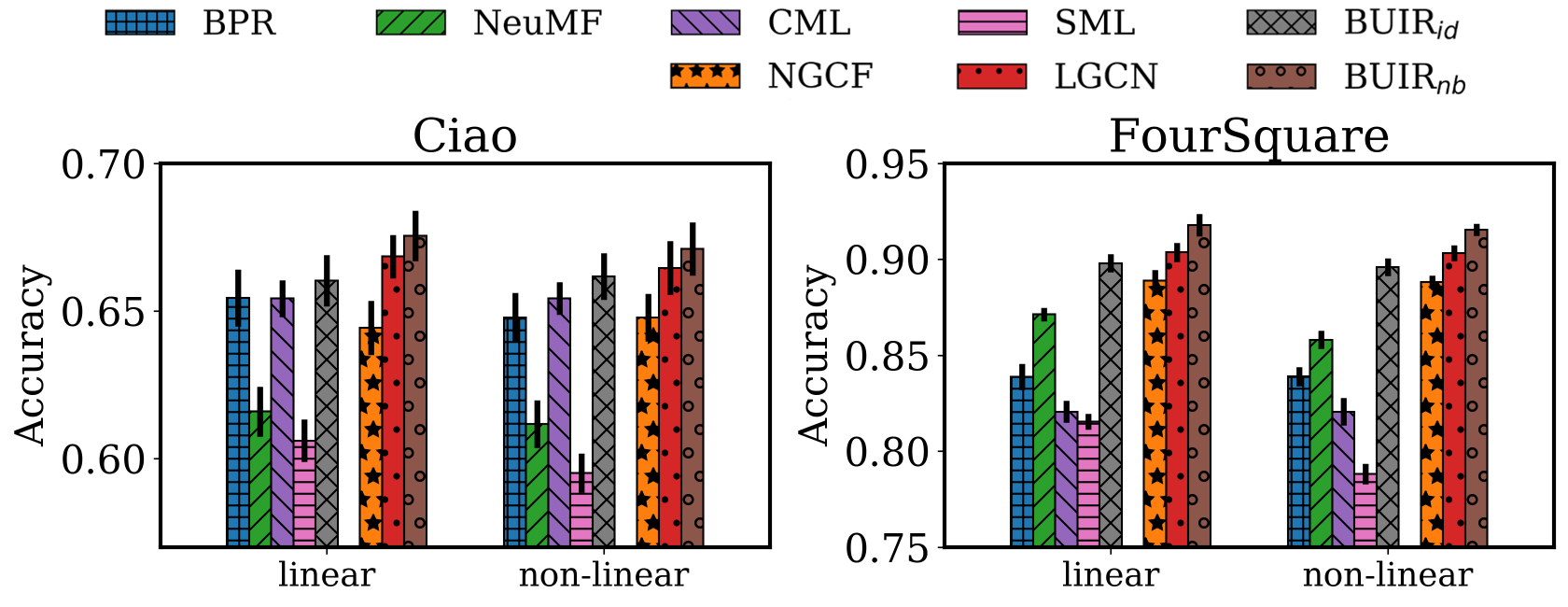}
	\caption{Evaluation on the quality of representations, by using a linear/non-linear classifier.} 
	\label{fig:downstream_perf}
\end{figure}

In Figure~\ref{fig:downstream_perf}, \proposedid and \proposednb achieve significantly higher classification accuracy than the others in each category.
This shows that the latent space induced by \proposed more accurately captures the item's characteristics (or their relationship) compared to the space induced by the baseline methods.
Another observation is that the rank of each method for the downstream tasks is consistent with that for top-$K$ recommendation (in Table~\ref{tbl:mainresults}).
It implies that the observed user-item interactions are positively-correlated with the latent semantic of the items, for this reason, effectively learning the users' implicit feedback eventually results in a good performance in the downstream tasks as well. 

\subsection{Sensitivity Analysis}
\label{subsec:exp_sensitivity}

For the guidance of hyperparameter selection, we provide analyses on the sensitivity of \proposed to its several hyperparameters.
We investigate the performance changes of \proposedid on the \foursq dataset ($\beta$=50\%) with respect to the dimension size $D$, the momentum coefficient $\tau$,\footnote{Considering that the target encoder should be slowly approximate the online encoder, we investigate $\tau$ in the range of [0.9, 1.0], as done in previous work~\cite{he2020momentum, grill2020bootstrap}.} and the number of layers in the predictor.

\begin{figure}[t]
	\centering
    \includegraphics[width=\linewidth]{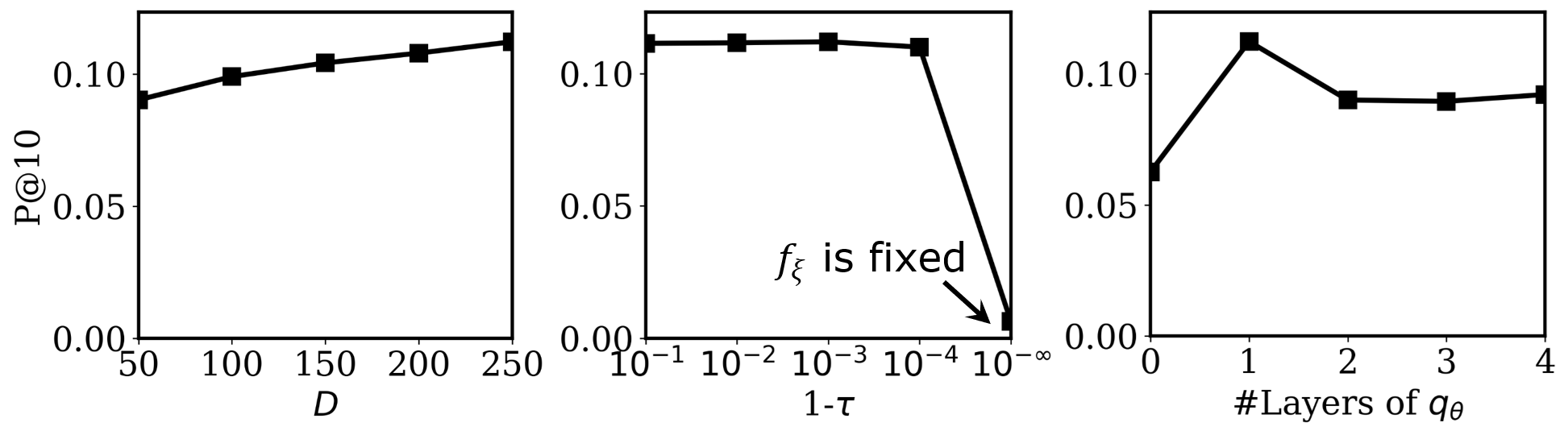}
	\caption{Sensitivity analyses on the \proposed hyperparameters.} 
	\label{fig:sensitivity_effect}
\end{figure}

Figure~\ref{fig:sensitivity_effect} clearly shows that the performance is hardly affected by $\tau$ in the range of [0.9, 1.0).
In other words, any values of $\tau$ larger than 0.9 allow the target encoder to successfully provide the target representations to the online encoder, by slowly approximating the online encoder;
on the contrary, \proposed cannot learn the effective representations at all in case that the target encoder is fixed (i.e., $\tau=1$).
This observation is consistent with previous work on momentum-based moving average~\cite{tarvainen2017mean, he2020momentum, grill2020bootstrap} that showed all values of $\tau$ between 0.9 and 0.999 can yield the best performance.
Furthermore, \proposed performs the best with a single-layer predictor, because a multi-layer predictor makes it difficult to optimize the relationship between outputs of the two encoder networks.
In conclusion, \proposed is more powerful even with fewer hyperparameters, compared to existing OCCF methods that include a variety of regularization terms or modeling components.

\section{Conclusion}
\label{sec:conc}
This paper proposes a novel framework for learning the representations of users and items, termed as \proposed, to address the main challenges of the OCCF problem: the implicit assumption about negative interactions, and high sparsity of observed (positively-labeled) interactions.
First, \proposed directly bootstraps the representations of users and items by minimizing their cross-prediction error.
This makes \proposed use only partially-observed positive interactions for training the model, and accordingly, it can eliminate the need for negative sampling.
In addition, \proposed is able to learn the augmented views of each positive interaction obtained from the neighborhood information, which further relieves the data sparsity issue of the OCCF problem.
Through the extensive comparison with a wide range of OCCF methods, we demonstrate that \proposed consistently outperforms all the other baselines in terms of top-$K$ recommendation.
In particular, the effectiveness of \proposed becomes more significant for much sparse datasets in which the positively-labeled interactions are not enough to optimize the model as well as the assumption about negative interactions becomes less valid. 
Based on its great compatibility with existing user/item encoder networks, we expect that our \proposed framework can be a major solution for the OCCF problem, replacing the conventional BPR framework. 

\begin{acks}
This work was supported by the NRF grant funded by the MSIT (No.
2020R1A2B5B03097210, 2021R1C1C1009081), and the IITP grant funded by the MSIT
(No. 2018-0-00584, 2019-0-01906).
\end{acks}

\bibliographystyle{ACM-Reference-Format}
\balance
\bibliography{BIB/bibliography}


\begin{thebibliography}{33}


\ifx \showCODEN    \undefined \def \showCODEN     #1{\unskip}     \fi
\ifx \showDOI      \undefined \def \showDOI       #1{#1}\fi
\ifx \showISBNx    \undefined \def \showISBNx     #1{\unskip}     \fi
\ifx \showISBNxiii \undefined \def \showISBNxiii  #1{\unskip}     \fi
\ifx \showISSN     \undefined \def \showISSN      #1{\unskip}     \fi
\ifx \showLCCN     \undefined \def \showLCCN      #1{\unskip}     \fi
\ifx \shownote     \undefined \def \shownote      #1{#1}          \fi
\ifx \showarticletitle \undefined \def \showarticletitle #1{#1}   \fi
\ifx \showURL      \undefined \def \showURL       {\relax}        \fi
\providecommand\bibfield[2]{#2}
\providecommand\bibinfo[2]{#2}
\providecommand\natexlab[1]{#1}
\providecommand\showeprint[2][]{arXiv:#2}

\bibitem[\protect\citeauthoryear{Chae, Kang, Kim, and Lee}{Chae
  et~al\mbox{.}}{2018}]%
        {chae2018cfgan}
\bibfield{author}{\bibinfo{person}{Dong-Kyu Chae}, \bibinfo{person}{Jin-Soo
  Kang}, \bibinfo{person}{Sang-Wook Kim}, {and} \bibinfo{person}{Jung-Tae
  Lee}.} \bibinfo{year}{2018}\natexlab{}.
\newblock \showarticletitle{Cfgan: A generic collaborative filtering framework
  based on generative adversarial networks}. In
  \bibinfo{booktitle}{\emph{CIKM}}. \bibinfo{pages}{137--146}.
\newblock


\bibitem[\protect\citeauthoryear{Chen, Kornblith, Norouzi, and Hinton}{Chen
  et~al\mbox{.}}{2020}]%
        {chen2020simple}
\bibfield{author}{\bibinfo{person}{Ting Chen}, \bibinfo{person}{Simon
  Kornblith}, \bibinfo{person}{Mohammad Norouzi}, {and}
  \bibinfo{person}{Geoffrey Hinton}.} \bibinfo{year}{2020}\natexlab{}.
\newblock \showarticletitle{A simple framework for contrastive learning of
  visual representations}. In \bibinfo{booktitle}{\emph{ICML}}.
\newblock


\bibitem[\protect\citeauthoryear{Chen and He}{Chen and He}{2021}]%
        {chen2020exploring}
\bibfield{author}{\bibinfo{person}{Xinlei Chen} {and} \bibinfo{person}{Kaiming
  He}.} \bibinfo{year}{2021}\natexlab{}.
\newblock \showarticletitle{Exploring Simple Siamese Representation Learning}.
  In \bibinfo{booktitle}{\emph{CVPR}}.
\newblock


\bibitem[\protect\citeauthoryear{Devlin, Chang, Lee, and Toutanova}{Devlin
  et~al\mbox{.}}{2019}]%
        {devlin2019bert}
\bibfield{author}{\bibinfo{person}{Jacob Devlin}, \bibinfo{person}{Ming-Wei
  Chang}, \bibinfo{person}{Kenton Lee}, {and} \bibinfo{person}{Kristina
  Toutanova}.} \bibinfo{year}{2019}\natexlab{}.
\newblock \showarticletitle{{BERT}: Pre-training of Deep Bidirectional
  Transformers for Language Understanding}. In
  \bibinfo{booktitle}{\emph{NAACL}}. \bibinfo{pages}{4171--4186}.
\newblock


\bibitem[\protect\citeauthoryear{Ding, Yu, He, Feng, Li, and Jin}{Ding
  et~al\mbox{.}}{2019}]%
        {ding2019sampler}
\bibfield{author}{\bibinfo{person}{Jingtao Ding}, \bibinfo{person}{Guanghui
  Yu}, \bibinfo{person}{Xiangnan He}, \bibinfo{person}{Fuli Feng},
  \bibinfo{person}{Yong Li}, {and} \bibinfo{person}{Depeng Jin}.}
  \bibinfo{year}{2019}\natexlab{}.
\newblock \showarticletitle{Sampler design for bayesian personalized ranking by
  leveraging view data}.
\newblock \bibinfo{journal}{\emph{TKDE}} (\bibinfo{year}{2019}).
\newblock


\bibitem[\protect\citeauthoryear{Grill, Strub, Altch{\'e}, Tallec, Richemond,
  Buchatskaya, Doersch, Pires, Guo, Azar, Piot, Kavukcuoglu, Munos, and
  Valko}{Grill et~al\mbox{.}}{2020}]%
        {grill2020bootstrap}
\bibfield{author}{\bibinfo{person}{Jean-Bastien Grill},
  \bibinfo{person}{Florian Strub}, \bibinfo{person}{Florent Altch{\'e}},
  \bibinfo{person}{Corentin Tallec}, \bibinfo{person}{Pierre~H Richemond},
  \bibinfo{person}{Elena Buchatskaya}, \bibinfo{person}{Carl Doersch},
  \bibinfo{person}{Bernardo~Avila Pires}, \bibinfo{person}{Zhaohan~Daniel Guo},
  \bibinfo{person}{Mohammad~Gheshlaghi Azar}, \bibinfo{person}{Bilal Piot},
  \bibinfo{person}{Koray Kavukcuoglu}, \bibinfo{person}{Rémi Munos}, {and}
  \bibinfo{person}{Michal Valko}.} \bibinfo{year}{2020}\natexlab{}.
\newblock \showarticletitle{Bootstrap your own latent: A new approach to
  self-supervised learning}. In \bibinfo{booktitle}{\emph{NeurIPS}}.
  \bibinfo{pages}{21271--21284}.
\newblock


\bibitem[\protect\citeauthoryear{Gutmann and Hyv{\"a}rinen}{Gutmann and
  Hyv{\"a}rinen}{2010}]%
        {gutmann2010noise}
\bibfield{author}{\bibinfo{person}{Michael Gutmann} {and} \bibinfo{person}{Aapo
  Hyv{\"a}rinen}.} \bibinfo{year}{2010}\natexlab{}.
\newblock \showarticletitle{Noise-contrastive estimation: A new estimation
  principle for unnormalized statistical models}. In
  \bibinfo{booktitle}{\emph{AISTATS}}.
\newblock


\bibitem[\protect\citeauthoryear{He, Fan, Wu, Xie, and Girshick}{He
  et~al\mbox{.}}{2020b}]%
        {he2020momentum}
\bibfield{author}{\bibinfo{person}{Kaiming He}, \bibinfo{person}{Haoqi Fan},
  \bibinfo{person}{Yuxin Wu}, \bibinfo{person}{Saining Xie}, {and}
  \bibinfo{person}{Ross Girshick}.} \bibinfo{year}{2020}\natexlab{b}.
\newblock \showarticletitle{Momentum contrast for unsupervised visual
  representation learning}. In \bibinfo{booktitle}{\emph{CVPR}}.
\newblock


\bibitem[\protect\citeauthoryear{He and McAuley}{He and McAuley}{2016}]%
        {he2016vbpr}
\bibfield{author}{\bibinfo{person}{Ruining He} {and} \bibinfo{person}{Julian
  McAuley}.} \bibinfo{year}{2016}\natexlab{}.
\newblock \showarticletitle{VBPR: visual Bayesian Personalized Ranking from
  implicit feedback}. In \bibinfo{booktitle}{\emph{AAAI}}.
  \bibinfo{pages}{144--150}.
\newblock


\bibitem[\protect\citeauthoryear{He, Deng, Wang, Li, Zhang, and Wang}{He
  et~al\mbox{.}}{2020a}]%
        {he2020lightgcn}
\bibfield{author}{\bibinfo{person}{Xiangnan He}, \bibinfo{person}{Kuan Deng},
  \bibinfo{person}{Xiang Wang}, \bibinfo{person}{Yan Li},
  \bibinfo{person}{Yongdong Zhang}, {and} \bibinfo{person}{Meng Wang}.}
  \bibinfo{year}{2020}\natexlab{a}.
\newblock \showarticletitle{Lightgcn: Simplifying and powering graph
  convolution network for recommendation}. In
  \bibinfo{booktitle}{\emph{SIGIR}}. \bibinfo{pages}{639--648}.
\newblock


\bibitem[\protect\citeauthoryear{He, Liao, Zhang, Nie, Hu, and Chua}{He
  et~al\mbox{.}}{2017}]%
        {he2017neural}
\bibfield{author}{\bibinfo{person}{Xiangnan He}, \bibinfo{person}{Lizi Liao},
  \bibinfo{person}{Hanwang Zhang}, \bibinfo{person}{Liqiang Nie},
  \bibinfo{person}{Xia Hu}, {and} \bibinfo{person}{Tat-Seng Chua}.}
  \bibinfo{year}{2017}\natexlab{}.
\newblock \showarticletitle{Neural collaborative filtering}. In
  \bibinfo{booktitle}{\emph{WWW}}. \bibinfo{pages}{173--182}.
\newblock


\bibitem[\protect\citeauthoryear{Hsieh, Yang, Cui, Lin, Belongie, and
  Estrin}{Hsieh et~al\mbox{.}}{2017}]%
        {hsieh2017collaborative}
\bibfield{author}{\bibinfo{person}{Cheng-Kang Hsieh}, \bibinfo{person}{Longqi
  Yang}, \bibinfo{person}{Yin Cui}, \bibinfo{person}{Tsung-Yi Lin},
  \bibinfo{person}{Serge Belongie}, {and} \bibinfo{person}{Deborah Estrin}.}
  \bibinfo{year}{2017}\natexlab{}.
\newblock \showarticletitle{Collaborative metric learning}. In
  \bibinfo{booktitle}{\emph{WWW}}. \bibinfo{pages}{193--201}.
\newblock


\bibitem[\protect\citeauthoryear{Hu, Koren, and Volinsky}{Hu
  et~al\mbox{.}}{2008}]%
        {hu2008collaborative}
\bibfield{author}{\bibinfo{person}{Yifan Hu}, \bibinfo{person}{Yehuda Koren},
  {and} \bibinfo{person}{Chris Volinsky}.} \bibinfo{year}{2008}\natexlab{}.
\newblock \showarticletitle{Collaborative filtering for implicit feedback
  datasets}. In \bibinfo{booktitle}{\emph{ICDM}}. \bibinfo{pages}{263--272}.
\newblock


\bibitem[\protect\citeauthoryear{Kang, Hwang, Kweon, and Yu}{Kang
  et~al\mbox{.}}{2020}]%
        {kang2020rrd}
\bibfield{author}{\bibinfo{person}{SeongKu Kang}, \bibinfo{person}{Junyoung
  Hwang}, \bibinfo{person}{Wonbin Kweon}, {and} \bibinfo{person}{Hwanjo Yu}.}
  \bibinfo{year}{2020}\natexlab{}.
\newblock \showarticletitle{DE-RRD: A Knowledge Distillation Framework for
  Recommender System}. In \bibinfo{booktitle}{\emph{CIKM}}.
  \bibinfo{pages}{605--614}.
\newblock


\bibitem[\protect\citeauthoryear{Kim, Lee, and Shim}{Kim et~al\mbox{.}}{2019}]%
        {kim2019dual}
\bibfield{author}{\bibinfo{person}{Seunghyeon Kim}, \bibinfo{person}{Jongwuk
  Lee}, {and} \bibinfo{person}{Hyunjung Shim}.}
  \bibinfo{year}{2019}\natexlab{}.
\newblock \showarticletitle{Dual neural personalized ranking}. In
  \bibinfo{booktitle}{\emph{WWW}}. \bibinfo{pages}{863--873}.
\newblock


\bibitem[\protect\citeauthoryear{Krichene and Rendle}{Krichene and
  Rendle}{2020}]%
        {krichene2020sampled}
\bibfield{author}{\bibinfo{person}{Walid Krichene} {and}
  \bibinfo{person}{Steffen Rendle}.} \bibinfo{year}{2020}\natexlab{}.
\newblock \showarticletitle{On Sampled Metrics for Item Recommendation}. In
  \bibinfo{booktitle}{\emph{KDD}}. \bibinfo{pages}{1748--1757}.
\newblock


\bibitem[\protect\citeauthoryear{Li, Zhang, Zhu, Qian, Zang, Han, and Hu}{Li
  et~al\mbox{.}}{2020}]%
        {li2020symmetric}
\bibfield{author}{\bibinfo{person}{Mingming Li}, \bibinfo{person}{Shuai Zhang},
  \bibinfo{person}{Fuqing Zhu}, \bibinfo{person}{Wanhui Qian},
  \bibinfo{person}{Liangjun Zang}, \bibinfo{person}{Jizhong Han}, {and}
  \bibinfo{person}{Songlin Hu}.} \bibinfo{year}{2020}\natexlab{}.
\newblock \showarticletitle{Symmetric Metric Learning with Adaptive Margin for
  Recommendation}. In \bibinfo{booktitle}{\emph{AAAI}}.
  \bibinfo{pages}{4634--4641}.
\newblock


\bibitem[\protect\citeauthoryear{Liang, Krishnan, Hoffman, and Jebara}{Liang
  et~al\mbox{.}}{2018}]%
        {liang2018variational}
\bibfield{author}{\bibinfo{person}{Dawen Liang}, \bibinfo{person}{Rahul~G
  Krishnan}, \bibinfo{person}{Matthew~D Hoffman}, {and} \bibinfo{person}{Tony
  Jebara}.} \bibinfo{year}{2018}\natexlab{}.
\newblock \showarticletitle{Variational autoencoders for collaborative
  filtering}. In \bibinfo{booktitle}{\emph{WWW}}. \bibinfo{pages}{689--698}.
\newblock


\bibitem[\protect\citeauthoryear{Liu, Wen, Jing, and Yu}{Liu
  et~al\mbox{.}}{2019}]%
        {liu2019deep}
\bibfield{author}{\bibinfo{person}{Huafeng Liu}, \bibinfo{person}{Jingxuan
  Wen}, \bibinfo{person}{Liping Jing}, {and} \bibinfo{person}{Jian Yu}.}
  \bibinfo{year}{2019}\natexlab{}.
\newblock \showarticletitle{Deep generative ranking for personalized
  recommendation}. In \bibinfo{booktitle}{\emph{RecSys}}.
  \bibinfo{pages}{34--42}.
\newblock


\bibitem[\protect\citeauthoryear{Liu, Pham, Cong, and Yuan}{Liu
  et~al\mbox{.}}{2017}]%
        {liu2017experimental}
\bibfield{author}{\bibinfo{person}{Yiding Liu},
  \bibinfo{person}{Tuan-Anh~Nguyen Pham}, \bibinfo{person}{Gao Cong}, {and}
  \bibinfo{person}{Quan Yuan}.} \bibinfo{year}{2017}\natexlab{}.
\newblock \showarticletitle{An Experimental Evaluation of Point-of-Interest
  Recommendation in Location-Based Social Networks}.
\newblock \bibinfo{journal}{\emph{PVLDB}} \bibinfo{volume}{10},
  \bibinfo{number}{10} (\bibinfo{date}{jun} \bibinfo{year}{2017}),
  \bibinfo{pages}{1010–1021}.
\newblock
\showISSN{2150-8097}


\bibitem[\protect\citeauthoryear{Mnih, Badia, Mirza, Graves, Lillicrap, Harley,
  Silver, and Kavukcuoglu}{Mnih et~al\mbox{.}}{2016}]%
        {mnih2016asynchronous}
\bibfield{author}{\bibinfo{person}{Volodymyr Mnih},
  \bibinfo{person}{Adria~Puigdomenech Badia}, \bibinfo{person}{Mehdi Mirza},
  \bibinfo{person}{Alex Graves}, \bibinfo{person}{Timothy Lillicrap},
  \bibinfo{person}{Tim Harley}, \bibinfo{person}{David Silver}, {and}
  \bibinfo{person}{Koray Kavukcuoglu}.} \bibinfo{year}{2016}\natexlab{}.
\newblock \showarticletitle{Asynchronous methods for deep reinforcement
  learning}. In \bibinfo{booktitle}{\emph{ICML}}. \bibinfo{pages}{1928--1937}.
\newblock


\bibitem[\protect\citeauthoryear{Mnih, Kavukcuoglu, Silver, Rusu, Veness,
  Bellemare, Graves, Riedmiller, Fidjeland, Ostrovski, et~al\mbox{.}}{Mnih
  et~al\mbox{.}}{2015}]%
        {mnih2015human}
\bibfield{author}{\bibinfo{person}{Volodymyr Mnih}, \bibinfo{person}{Koray
  Kavukcuoglu}, \bibinfo{person}{David Silver}, \bibinfo{person}{Andrei~A
  Rusu}, \bibinfo{person}{Joel Veness}, \bibinfo{person}{Marc~G Bellemare},
  \bibinfo{person}{Alex Graves}, \bibinfo{person}{Martin Riedmiller},
  \bibinfo{person}{Andreas~K Fidjeland}, \bibinfo{person}{Georg Ostrovski},
  {et~al\mbox{.}}} \bibinfo{year}{2015}\natexlab{}.
\newblock \showarticletitle{Human-level control through deep reinforcement
  learning}.
\newblock \bibinfo{journal}{\emph{nature}} \bibinfo{volume}{518},
  \bibinfo{number}{7540} (\bibinfo{year}{2015}), \bibinfo{pages}{529--533}.
\newblock


\bibitem[\protect\citeauthoryear{Oord, Li, and Vinyals}{Oord
  et~al\mbox{.}}{2018}]%
        {oord2018representation}
\bibfield{author}{\bibinfo{person}{Aaron van~den Oord}, \bibinfo{person}{Yazhe
  Li}, {and} \bibinfo{person}{Oriol Vinyals}.} \bibinfo{year}{2018}\natexlab{}.
\newblock \showarticletitle{Representation learning with contrastive predictive
  coding}.
\newblock \bibinfo{journal}{\emph{arXiv preprint arXiv:1807.03748}}
  (\bibinfo{year}{2018}).
\newblock


\bibitem[\protect\citeauthoryear{Pan, Zhou, Cao, Liu, Lukose, Scholz, and
  Yang}{Pan et~al\mbox{.}}{2008}]%
        {pan2008one}
\bibfield{author}{\bibinfo{person}{Rong Pan}, \bibinfo{person}{Yunhong Zhou},
  \bibinfo{person}{Bin Cao}, \bibinfo{person}{Nathan~N Liu},
  \bibinfo{person}{Rajan Lukose}, \bibinfo{person}{Martin Scholz}, {and}
  \bibinfo{person}{Qiang Yang}.} \bibinfo{year}{2008}\natexlab{}.
\newblock \showarticletitle{One-class collaborative filtering}. In
  \bibinfo{booktitle}{\emph{ICDM}}. \bibinfo{pages}{502--511}.
\newblock


\bibitem[\protect\citeauthoryear{Park, Kim, Xie, and Yu}{Park
  et~al\mbox{.}}{2018}]%
        {park2018collaborative}
\bibfield{author}{\bibinfo{person}{Chanyoung Park}, \bibinfo{person}{Donghyun
  Kim}, \bibinfo{person}{Xing Xie}, {and} \bibinfo{person}{Hwanjo Yu}.}
  \bibinfo{year}{2018}\natexlab{}.
\newblock \showarticletitle{Collaborative translational metric learning}. In
  \bibinfo{booktitle}{\emph{ICDM}}. \bibinfo{pages}{367--376}.
\newblock


\bibitem[\protect\citeauthoryear{Rendle and Freudenthaler}{Rendle and
  Freudenthaler}{2014}]%
        {rendle2014improving}
\bibfield{author}{\bibinfo{person}{Steffen Rendle} {and}
  \bibinfo{person}{Christoph Freudenthaler}.} \bibinfo{year}{2014}\natexlab{}.
\newblock \showarticletitle{Improving pairwise learning for item recommendation
  from implicit feedback}. In \bibinfo{booktitle}{\emph{WSDM}}.
  \bibinfo{pages}{273--282}.
\newblock


\bibitem[\protect\citeauthoryear{Rendle, Freudenthaler, Gantner, and
  Schmidt-Thieme}{Rendle et~al\mbox{.}}{2009}]%
        {rendle2009bpr}
\bibfield{author}{\bibinfo{person}{Steffen Rendle}, \bibinfo{person}{Christoph
  Freudenthaler}, \bibinfo{person}{Zeno Gantner}, {and} \bibinfo{person}{Lars
  Schmidt-Thieme}.} \bibinfo{year}{2009}\natexlab{}.
\newblock \showarticletitle{BPR: Bayesian personalized ranking from implicit
  feedback}. In \bibinfo{booktitle}{\emph{UAI}}.
\newblock


\bibitem[\protect\citeauthoryear{Tang, Gao, and Liu}{Tang
  et~al\mbox{.}}{2012}]%
        {tang2012mtrust}
\bibfield{author}{\bibinfo{person}{Jiliang Tang}, \bibinfo{person}{Huiji Gao},
  {and} \bibinfo{person}{Huan Liu}.} \bibinfo{year}{2012}\natexlab{}.
\newblock \showarticletitle{mTrust: discerning multi-faceted trust in a
  connected world}. In \bibinfo{booktitle}{\emph{WSDM}}.
  \bibinfo{pages}{93--102}.
\newblock


\bibitem[\protect\citeauthoryear{Tarvainen and Valpola}{Tarvainen and
  Valpola}{2017}]%
        {tarvainen2017mean}
\bibfield{author}{\bibinfo{person}{Antti Tarvainen} {and}
  \bibinfo{person}{Harri Valpola}.} \bibinfo{year}{2017}\natexlab{}.
\newblock \showarticletitle{Mean teachers are better role models:
  Weight-averaged consistency targets improve semi-supervised deep learning
  results}. In \bibinfo{booktitle}{\emph{NeurIPS}}.
  \bibinfo{pages}{1195--1204}.
\newblock


\bibitem[\protect\citeauthoryear{Wang, Chen, and Li}{Wang
  et~al\mbox{.}}{2013}]%
        {wang2013collaborative}
\bibfield{author}{\bibinfo{person}{Hao Wang}, \bibinfo{person}{Binyi Chen},
  {and} \bibinfo{person}{Wu-Jun Li}.} \bibinfo{year}{2013}\natexlab{}.
\newblock \showarticletitle{Collaborative topic regression with social
  regularization for tag recommendation}. In \bibinfo{booktitle}{\emph{IJCAI}}.
\newblock


\bibitem[\protect\citeauthoryear{Wang, Yu, Zhang, Gong, Xu, Wang, Zhang, and
  Zhang}{Wang et~al\mbox{.}}{2017}]%
        {wang2017irgan}
\bibfield{author}{\bibinfo{person}{Jun Wang}, \bibinfo{person}{Lantao Yu},
  \bibinfo{person}{Weinan Zhang}, \bibinfo{person}{Yu Gong},
  \bibinfo{person}{Yinghui Xu}, \bibinfo{person}{Benyou Wang},
  \bibinfo{person}{Peng Zhang}, {and} \bibinfo{person}{Dell Zhang}.}
  \bibinfo{year}{2017}\natexlab{}.
\newblock \showarticletitle{Irgan: A minimax game for unifying generative and
  discriminative information retrieval models}. In
  \bibinfo{booktitle}{\emph{SIGIR}}. \bibinfo{pages}{515--524}.
\newblock


\bibitem[\protect\citeauthoryear{Wang, He, Wang, Feng, and Chua}{Wang
  et~al\mbox{.}}{2019}]%
        {wang2019neural}
\bibfield{author}{\bibinfo{person}{Xiang Wang}, \bibinfo{person}{Xiangnan He},
  \bibinfo{person}{Meng Wang}, \bibinfo{person}{Fuli Feng}, {and}
  \bibinfo{person}{Tat-Seng Chua}.} \bibinfo{year}{2019}\natexlab{}.
\newblock \showarticletitle{Neural graph collaborative filtering}. In
  \bibinfo{booktitle}{\emph{SIGIR}}. \bibinfo{pages}{165--174}.
\newblock


\bibitem[\protect\citeauthoryear{Wu, DuBois, Zheng, and Ester}{Wu
  et~al\mbox{.}}{2016}]%
        {wu2016collaborative}
\bibfield{author}{\bibinfo{person}{Yao Wu}, \bibinfo{person}{Christopher
  DuBois}, \bibinfo{person}{Alice~X Zheng}, {and} \bibinfo{person}{Martin
  Ester}.} \bibinfo{year}{2016}\natexlab{}.
\newblock \showarticletitle{Collaborative denoising auto-encoders for top-n
  recommender systems}. In \bibinfo{booktitle}{\emph{WSDM}}.
\newblock


\end{thebibliography}


\end{document}